\documentclass{article}

\usepackage{arxiv}

\usepackage[utf8]{inputenc} 
\usepackage[T1]{fontenc}    
\usepackage{hyperref}       
\usepackage{url}            
\usepackage{booktabs}       
\usepackage{amsfonts}       
\usepackage{nicefrac}       
\usepackage{microtype}      
\usepackage{lipsum}		
\usepackage{graphicx}
\usepackage{natbib}
\usepackage{doi}
\usepackage{amsmath}
\usepackage{amssymb}
\usepackage{mathrsfs}
\newtheorem{theorem}{Theorem}
\newtheorem{lemma}[theorem]{Lemma} 
\newtheorem{proposition}[theorem]{Proposition} 

\newtheorem{corollary}[theorem]{Corollary}
\newtheorem{definition}[theorem]{Definition}

\usepackage{diagbox}
\usepackage{multirow}
\usepackage{graphicx}  
\usepackage{float}  
\usepackage{subfigure}  

\newcommand{\BlackBox}{\rule{1.5ex}{1.5ex}}  

\title{Comprehensive Analysis of Over-smoothing in Graph Neural Networks from Markov Chains Perspective}

\author{ Weichen Zhao, Chenguang Wang, Congying Han\thanks{Corresponding author at: No.19A, Yuquan Road, Shijingshan District, Beijing.} \& Tiande Guo \\
School of Mathematical Sciences\\
University of Chinese Academy of Sciences (UCAS)\\
Key Laboratory of Big Data Mining and Knowledge Management, CAS\\
		Beijing, China \\
\texttt{\{zhaoweichen14, wangchenguang19\}@mails.ucas.ac.cn,\{hancy, tdguo\}@ucas.ac.cn} \\
}

\renewcommand{\shorttitle}{\textit{arXiv} Template}

\begin{document}
\maketitle

\begin{abstract}
	The over-smoothing problem is an obstacle of developing deep graph neural network (GNN). Although many approaches to improve the over-smoothing problem have been proposed, there is still a lack of comprehensive understanding and conclusion of this problem. In this work, we analyze the over-smoothing problem from the Markov chain perspective. We focus on message passing of GNN and first establish a connection between GNNs and Markov chains on the graph. GNNs are divided into two classes of operator-consistent and operator-inconsistent based on whether the corresponding Markov chains are time-homogeneous. Next we attribute the over-smoothing problem to the convergence of an arbitrary initial distribution to a stationary distribution. Based on this, we prove that although the previously proposed methods can alleviate over-smoothing, but these methods cannot avoid the over-smoothing problem. In addition, we give the conclusion of the over-smoothing problem in two types of GNNs in the Markovian sense. On the one hand, operator-consistent GNN cannot avoid over-smoothing at an exponential rate. On the other hand, operator-inconsistent GNN is not always over-smoothing. Further, we investigate the existence of the limiting distribution of the time-inhomogeneous Markov chain, from which we derive a sufficient condition for operator-inconsistent GNN to avoid over-smoothing. Finally, we design experiments to verify our findings. Results show that our proposed sufficient condition can effectively improve over-smoothing problem in operator-inconsistent GNN and enhance the performance of the model.
\end{abstract}


\section{Introduction}
Graph neural networks \cite{2016Semi,bruna2013spectral,defferrard2016convolutional,velivckovic2017graph,abu2018watch,zhang2018gaan,lee2018graph,klicpera2018predict} have achieved great success in processing graph data which is rich in information about the relationships between objects, and have been successfully applied to chemistry \cite{2019Graph,kearnes2016molecular,de2018molgan,gilmer2017neural}, traffic prediction \cite{ 2019Traffic,2019Predicting,2019Predicting2}, knowledge graph \cite{2019Estimating,2019Knowledge}, social network \cite{2019Learning,qiu2018deepinf}, recommendation system \cite{2018Graph} and other aspects. In recent years, the rapid development of graph neural network research has been accompanied by some problems that show more demand for mathematical explanations. Many scholars analyze GNNs using mathematical tools, including Weisfeiler-Lehman tests \cite{xu2018powerful,maron2019provably,azizian2020expressive}, spectral analysis \cite{nt2019revisiting,li2018deeper,wu2019simplifying} and dynamical system \cite{oono2019graph}. However, like other deep neural networks, theoretical analysis related to GNNs is still scarce. Different perspectives are needed to expand scholars' understanding and promote the development of GNNs.
	
	The deepening of the network has brought about changes in neural networks and caused a boom in deep learning. Unlike typical deep neural networks, in the training of graph neural networks, researchers have found that the performance of GNN decreases instead as the depth increases. There are several possible reasons for the depth limitations of GNN. Li et al. \cite{li2018deeper} first attribute this anomaly to \emph{over-smoothing}, a phenomenon in which the representations of different nodes tend to be consistent as the network deepens, leading to indistinguishable node representations. Many researchers have studied this problem and proposed some improvement methods\cite{li2018deeper,oono2019graph,rong2019dropedge,chen2020simple,chen2020measuring,huang2020tackling, cai2020note,yang2020revisiting,chiang2019cluster,li2020deepergcn}. However, there is still no unified framework to prove the effectiveness of these methods. In addition, theoretical analysis of related works only focus on specific models such as \emph{graph convolution network} (GCN) or \emph{graph attention network} (GAT), and lack a comprehensive analysis and understanding of the general graph neural network. 
	
	First, the over-smoothing problem in GNN needs to be modeled as a mathematical problem. Noting the Markov property of the forward propagation process of GNNs and considering the node set as a state space, in this work, we connect GNNs with Markov chains on the graph. By analogy with the time-homogeneousness of Markov chains, we divide GNNs into two categories of operator-consistent models and operator-inconsistent models based on whether the message passing operators of each layer are consistent. Specifically, we connect GCN and GAT with a simple random walk and a time-inhomogeneous random walk on the graph, respectively. In addition, we model the stochastic method, DropEdge \cite{rong2019dropedge}, as a random environment and establish a connection between the GNN models which use the DropEdge method and \emph{Markov Chains in Random Environments} (MCRE) \cite{cogburn1980markov,cogburn1990direct,nawrotzki1982finite,orey1991markov}. Considering the nodes' representations as distributions on the state space, we attribute the over-smoothing in GNN to the convergence of the representation distribution to the stationary distribution. 
	
	Based on this, we analyze previous methods for improving over-smoothing including residual connections method \cite{2016Semi,chiang2019cluster}, personalized propagation of neural predictions (PPNP) \cite{klicpera2018predict}, and the DropEdge method \cite{rong2019dropedge,huang2020tackling} from the perspective of Markov chains. By studying the lazy walk on the graph, we show that these methods can alleviate the over-smoothing problem. However, we prove that the lazy walk on the graph still has stationary distribution and the rate of convergence is exponential. This shows that these methods can not avoid the over-smoothing problem in GNN.
	
	Next, we give conclusions on whether the general GNN model can avoid the over-smoothing in the Markovian sense. By studying the existence of stationary distribution of the time-homogeneous chain, we state that for the operator-consistent GNN, the node features cannot avoid over-smoothing, nor can they avoid over-smoothing at the exponential rate. Using conclusions of \cite{bowerman1977convergence,huang1976rate} in the time-inhomogeneous Markov chain, we show that operator-inconsistent GNN does not necessarily suffer from over-smoothing. 
	
	Further, we try to solve the over-smoothing problem in operator-inconsistent GNN. In general, we prove a necessary condition for the existence of the stationary distribution of the time-inhomogeneous chain. Based on this conclusion, we take GAT which is the most typical operator-inconsistent GNN model as an example, derive a sufficient condition for GAT to avoid over-smoothing.
	
	Finally, we verify our conclusions on the benchmark datasets. Based on the sufficient condition, we propose a regularization term which can be flexibly added to the training of the neural network. Results show that our proposed sufficient condition can significantly improve the performance of model. In addition, the representation learned by different nodes is more inconsistent after adding the regularization term, which indicates that the over-smoothing in GAT is improved.\\
	
	\noindent
	\textbf{Contribution.} In summary, our contributions are as follows:
	\begin{itemize}
		\item We establish a connection between GNNs and Markov chains on the graph, implementing a mathematical framework for exploring the over-smoothing problem in GNNs (Section \ref{3}). 
		\item We reveal the cause of the over-smoothing problem, attributing it to the convergence of feature distribution to a stationary distribution (Section \ref{4.1}).
		\item We prove the effectiveness of previous methods to alleviate the over-smoothing, however, they cannot avoid over-smoothing from the perspective of message passing (Section \ref{4.2}).
		\item We give conclusions on whether the general GNN models are able to avoid the over-smoothing problem in the Markovian sense (Section \ref{4.3}).
		\item We study the existence of limiting distributions of the time-inhomogeneous Markov chain. And based on this, we give a sufficient condition for operator-inconsistent GNN to avoid over-smoothing (Section \ref{4.4}).
		\item We propose a regularization term based on this sufficient condition and experimentally verify that our proposed condition can improve the model performance by solving the over-smoothing problem from the perspective of message passing (Section \ref{5}).
	\end{itemize}
	
	\noindent
	\textbf{Notation.}
	Let $ \mathcal{G}=(\mathcal{V},\mathcal{E}) $ be a connected non-bipartite graph, where $ \mathcal{V}:=\{1,2,\ldots,N\} $ is the node set, $ \mathcal{E} $ is the edge set, $ N=|\mathcal{V}| $ is the number of nodes and $ |\mathcal{A}| $ denotes the number of elements in the set $ \mathcal{A} $. If there are connected edges between nodes $ u,v\in\mathcal{V} $, then denote by $ (u,v)\in\mathcal{E} $. $ \deg(u) $ denotes the degree of node $ u\in\mathcal{V} $ and $ \mathcal{N}(v) $ denotes the neighbors of node $ v $. The corresponding adjacency matrix is $ A $ and the degree matrix is $ D $. $ \tilde{\mathcal{G}}=(\tilde{\mathcal{V}},\tilde{\mathcal{E}}) $ denotes the graph $ \mathcal{G} $ added self-loop, the corresponding adjacency matrix is $ \tilde{A} $ and the degree matrix is $ \tilde{D} $.
	
	Let $ \mathbb{R} $ be the set of real numbers, $ \mathbb{N} $ be the set of natural numbers, $ \mathbb{Z}^{+} $ be the set of positive integers, and $ F $ be the features dimensions of the nodes. Let $ (\Omega,\mathcal{F},\mathbf{P}) $ be a probability space and $ \mathbf{E} $ be the expectation operator on it. $ \|\cdot\| $ denotes the $ L^{1} $ norm and $ \|\cdot\|_{TV} $ denotes the total variation distance. $ P^{\text{T}} $ denotes the transposed matrix of the square matrix $ P $. $ p(i,j) $ denotes the $ i $th row $ j $th column element of the matrix. $ P(i,\cdot) $ denotes the row vector formed by the $ i $th row of the matrix $ P $ and $ P(\cdot,j) $ denotes the column vector formed by the $ j $th column of the matrix $ P $.
	
	Let $ P_{\text{rw}}:=D^{-1}A $ be the transition matrix of simple random walk on the graph $ \mathcal{G} $, $ \tilde{P}_{\text{rw}}:=\tilde{D}^{-1}\tilde{A} $ be the transition matrix of simple random walk on the graph $ \tilde{\mathcal{G}} $, and let $ P_{\text{lazy}} := (1-\gamma)D^{-1}A+\gamma I $ for the transition matrix of the lazy walk on the graph $ \mathcal{G} $.
\section{Preliminaries}\label{2}
	In this section we introduce the basic GNN model, the over-smoothing problem, and some definitions and conclusions in Markov chains theory.
	\subsection{Graph Neural Networks}\label{2.1}
	\textbf{Graph convolution networks} are the most important class of GNN models at present. Researchers have been working to migrate the success of convolutional neural networks to the learning of graph data. Given a graph $ \mathcal{G}=(\mathcal{V},\mathcal{E}) $ and the initial features of the nodes on it $ H^{(0)}\in\mathbb{R}^{N\times F} $, Kipf and Welling \cite{2016Semi} improved neural network models on graphs \cite{bruna2013spectral,defferrard2016convolutional} and proposed the most widely studied and applied vanilla GCN
	$$ H^{(l)}=\sigma_{W^{(l)}}\left(P_{\text{GCN}}H^{(l-1)}\right), $$
	where $ H^{(l)}\in\mathbb{R}^{N\times F} $ is the node feature vector output by the $ l $th hidden layer. $ W^{(l)} $ is the parameter of the $ l $th hidden layer. $ P_{\text{GCN}}:=\tilde{D}^{-\frac{1}{2}}\tilde{A}\tilde{D}^{-\frac{1}{2}} $ is the \emph{graph convolution operator}.
	
	\noindent
	\textbf{Graph attention networks.} Inspired by the attention mechanism, many scholars have proposed attention-based graph neural network models \cite{velivckovic2017graph,abu2018watch,zhang2018gaan,lee2018graph}. Among them, GAT \cite{velivckovic2017graph} is the most representative model. GAT establishes attention functions between nodes $ u,v\in\mathcal{E} $ with connected edges $ (u,v)\in\mathcal{V} $
	$$ \alpha_{u,v}^{(l)}=\frac{\exp(\phi^{(l)}(h_{u}^{(l-1)},h_{v}^{(l-1)}))}{\sum_{k\in\mathcal{N}(u)}\exp(\phi^{(l)}(h_{u}^{(l-1)},h_{k}^{(l-1)}))}, $$
	where $ h_{u}^{(l)}\in\mathbb{R}^{F} $ is the embedding for node $ u $ at the $ l $-th layer and
	$ \phi^{(l)}(h_{u}^{(l-1)},h_{v}^{(l-1)}):=\text{LeakyReLU}(\mathbf{a}^{\text{T}}[W^{(l)}h_{u}^{(l-1)}\|W^{(l)}h_{v}^{(l-1)}]), $
	where $ \mathbf{a}\in\mathbb{R}^{2F} $ and $ W^{(l)} $ is the weight matrix. Then the GAT layer is defined as
	$$ h_{u}^{(l)}:=\sigma_{W^{(l)}}(\sum_{v\in\mathcal{N}(u)}\alpha_{u,v}^{(l)}h_{v}^{(l-1)}). $$
	Written in matrix form
	$$ H^{(l)}=\sigma_{W^{(l)}}(P_{\text{att}}^{(l)}H^{(l-1)}), $$
	where $ P_{\text{att}}^{(l)}\in\mathbb{R}^{N\times N}$ is the attention matrix satisfying $ p_{\text{att}}^{(l)}(u,v)=\alpha_{u,v}^{(l)} $ if $ v\in\mathcal{N }(u) $ otherwise $ p_{\text{att}}^{(l)}(u,v)=0 $ and $ \sum_{v=1}^{N}p_{\text{att}}^{(l)}(u,v)=1 $.
	
	\noindent
	\textbf{Message Passing Neural Network} (MPNN) is a GNN model proposed by \cite{gilmer2017neural}. MPNN is a general framework for current GNN models, and most GNN models can be unified under this framework. It describes the GNN uniformly as a process that the information of the neighbors of a node $ u\in\mathcal{V} $ in a graph is passed as messages along an edge and aggregated at the node $ u $, i.e.
	$$ h_{u}^{(l)}=\text{UP}^{(l)}\left(h_{u}^{(l-1)},\text{MSG}^{(l)}(h_{v}^{(l-1)},v\in\mathcal{N}(u))\right), \eqno{(1)}$$ where $ h_{u}^{(l)} $ denotes the features of node $ u $ output at the $ l $th hidden layer, $ \text{UP}^{(l)} $ and $ \text{MSG}^{(l)} $ denote the update function and message passing function of the $ l $th layer, respectively. GNN such as GCN and GAT can be written in this form, differing only in the update function and message passing function.
	\subsection{Over-smoothing}\label{2.2}
	There is a phenomenon that the GNN has better experimental results in the shallow layer case, and instead do not work well in the deep layer case. The researchers find that this is due to the fact that during the GNN training process, the hidden layer representation of each node tend to converge to the same value as the number of layers increases. This phenomenon is called over-smoothing. This problem affects the deepening of GNN layers and limits the further development of GNN. The main current methods to alleviate over-smoothing are residual connections method \cite{2016Semi,chiang2019cluster}, PPNP \cite{klicpera2018predict}, and the DropEdge method \cite{rong2019dropedge,huang2020tackling}.
	
	\noindent
	\textbf{Residual connections method} is proposed based on an intuitive analysis of the over-smoothing problem. From a qualitative perspective, the over-smoothing problem is that as the network is stacked, the model forgets the initial input features and only updates the representations based on the structure of the graph data. It is natural to think that the problem of the model forgetting the initial features can be alleviated by reminding the network what its previous features are. Many methods have been proposed based on such intuitive analysis.  The simplest one, \cite{2016Semi} proposes to add residual connections to graph convolutional networks
	$$ H^{(l)}=\sigma_{W^{(l)}}\left(P_{\text{GCN}}H^{(l-1)}\right)+H^{(l-1)}. \eqno{(2)}$$
	The node features of the $ l \;$th hidden layer are directly added to the node features of the previous layer $ H^{(l-1)} $ to remind the network not to forget the previous features. However, \cite{chiang2019cluster} argues that residual connectivity ignores the structure of the graph and should be considered to reflect more the influence of the weights of different neighboring nodes. So this work gives more weight to the features from the previous layer in the message passing of each GCN layer by improving the graph convolution operator
	$$ H^{(l)}=\sigma_{W^{(l)}}\left(\left(P_{\text{GCN}}+I\right)H^{(l-1)}\right). \eqno{(3)}$$ \cite{chen2020simple,li2019deepgcns,xu2018representation} also use this idea.
	
	\noindent
	\textbf{Personalized propagation of neural predictions.} The well-known graph neural network PPNP \cite{klicpera2018predict} has also been experimentally proven to alleviate the over-smoothing. The innovation of PPNP is to decouple node information embedding and node feature propagation in GNN models. It simplifies the model by not learning the parameters during model propagation, and becomes one of the GNN models that are widely studied and applied \cite{bojchevski2020scaling}.
	
	PPNP's node information embedding is implemented by a MLP $ H^{(0)}=f_{W}(X). $
	Inspired by personalized PageRank (PPR), the node feature propagation process is
	$$ H^{(l)}=\sigma\left((1-\alpha)\tilde{D}^{-1}\tilde{A}H^{(l-1)}+\alpha H^{(0)}\right), \eqno{(4)}$$
	where $ \alpha\in(0,1) $ is the teleport probability. 
	
	\noindent
	\textbf{DropEdge.} \cite{oono2019graph} connects the GCN with a dynamical system and interprets the over-smoothing problem as the convergence of the dynamical system to an invariant subspace. \cite{rong2019dropedge} proposed DropEdge method based on the perspective of dynamical system. The idea of DropEdge is to randomly drop some edges in the original graph $ \tilde{\mathcal{G}}=(\tilde{\mathcal{V}},\tilde{\mathcal{E}}) $ at each layer. The specific operation is to randomly let some elements $ 1 $ of the adjacency matrix $ A $ become $ 0 $
	$$ A^{(l)}_{\text{drop}}=A-A'^{(l)}, $$
	where $ A'^{(l)} $ is the adjacency matrix formed by the expansion of a random subset $ \tilde{\mathcal{E}}' $ of $ \tilde{\mathcal{E}} $. This operation slows down the convergence of the dynamical system to the invariant subspace. Thus DropEdge method can alleviate the over-smoothing.
	
	Although these three methods have been experimentally verified to alleviate over-smoothing, their principles still lack explanation. Residual connections method only discusses the intuitive understanding, and PPNP is experimentally found to alleviate over-smoothing. DropEdge, as a widely used stochastic regularization method, needs more mathematical explanation from different perspective. In addition, the effectiveness of these three methods needs to be proved theoretically.
	
	\subsection{Results in Markov Chains}\label{2.3}
	Since the graph has a finite node set and the forward propagation process of graph neural networks is time-discrete, we focus on the results related to discrete-time Markov chains in finite state. This section introduces some conclusions of Markov chains, Markov chains in random environments \cite{cogburn1980markov,cogburn1990direct,nawrotzki1982finite} and mixing time \cite{levin2017markov}, which is an important tool to characterize the transitions of Markov chains. The results in this section are classical, which proofs can be found in books related to Markov chains.
	\begin{lemma}[Dobrushin's inequality]\label{lem2}
		Let $ \mu $ and $ \nu $ be probability distributions on a finite state space $ E $ and $ P $ be a transition matrix, then
		$$ \|\mu P-\nu P \|\leq  C(P)\|\mu-\nu\|, $$
		where $ C(P):=\frac{1}{2}\sup_{i,j}\sum_{k\in E}|p(i,k)-p(j,k)| $ is called the  Dobrushin contraction coefficient of the transition matrix $ P $.
	\end{lemma}
	\begin{lemma}\label{lem3}
		If one of the following two conditions is satisfied
		\begin{itemize}
			\item[(1)] $ P $ is irreducible and aperiodic.
			\item[(2)] $ C(P)<1. $
		\end{itemize}
		Then there exist stationary distribution $ \pi $, constants $ \alpha\in(0,1) $ and $ C>0 $ such that
		$$ \max_{i\in E}\| P^{n}(i,\;\cdot\;)-\pi\|\leq C \alpha^{n}. $$
	\end{lemma}
	Compared to the time-homogeneous Markov chain, it is much more difficult to investigate the limiting distribution of the time-inhomogeneous chain. \cite{bowerman1977convergence,huang1976rate} discussed the limiting case that an arbitrary initial distribution transfer according to a time-inhomogeneous chain. The sufficient condition for the existence of the limiting distribution is summarized in the following theorem.
	\begin{lemma}[Dobrushin-Isaacson-Madsen]\label{lem4}
		Let $ \vec{X}=\{X_{n},n\in T\} $ be a time-inhomogeneous Markov chain on a finite state space $ E $ with transition matrix $ P^{(n)} $. If the following $ (1), (2) $ and $ (3A) $ or $ (3B) $ are satisfied
		\begin{itemize}
			\item[(1)] There exists a stationary distribution $ \pi^{(n)} $ when $ P^{(n)} $ is treated as a transition matrix of a time-homogeneous chain;
			\item[(2)] $ \sum_{n}\|\pi^{(n)}-\pi^{(n+1)}\|<\infty; $
			\item[(3A)] (Isaacson-Madsen condition) For any probability distribution $ \mu $, $ \nu $ on $ E $ and positive integer $ k $
			$$ \|(\mu-\nu)P^{(k)}\cdots P^{(n)}\|\rightarrow 0,\quad n\rightarrow \infty. $$
			\item[(3B)] (Dobrushin condition) For any positive integers $ k $
			$$ C(P^{(k)}\cdots P^{(n)})\rightarrow 0,\quad n\rightarrow \infty. $$
			
		\end{itemize}
		Then there exists a probability measure $ \pi $ on $ E $ such that
		\begin{itemize}
			\item[(1)] $ \|\pi^{(n)}-\pi \|\rightarrow 0,\quad n\rightarrow \infty; $
			\item[(2)] Let the initial distribution be $ \mu_{0} $ and the distribution of the chain $ \vec{X} $ at step $ n $ be $ \mu_{n}:=\mu_{n-1}P^{(n)} $, then for any initial distribution $ \mu_{0} $, we have $ \|\mu_{n}-\pi \|\rightarrow 0,\; n\rightarrow \infty. $
		\end{itemize}
	\end{lemma}
	
	Although Markov chains are widely used in real-world engineering, some practical problems have more complex environmental factors that affect the transition of the original Markov chain. We need a more advanced mathematical tool to deal with such problems. If these complex environmental factors are described as a stochastic process that affects the transition function of the original chain, the original chain may lose the Markov property as a result, so researchers \cite{cogburn1980markov,cogburn1990direct,nawrotzki1982finite,orey1991markov} developed a theory of \emph{Markov chains in random environments} (MCRE) to deal with these problems.
	
	Let $ (E,\mathcal{E}) $, $ (Y,\mathcal{Y}) $ be two state spaces, $ T_{1}=\mathbb{N}=\{0,1,2,\ldots\} $, $ T_{2}=\mathbb{Z}^{+}=\{1,2,\ldots\} $ are two sets of times. In the following we introduce the random Markov kernel that couples the original chain to the random environment and a Markov chain in a positive time-dependent random environment with finite state space.
	
	\begin{definition}[random Markov kernel]
		Let
		$ p(\cdot;\cdot,\cdot):Y\times E\times\mathcal{E}\mapsto[0,1] $. If 
		\begin{itemize}
			\item[(1)] For every fixed $ \theta\in Y $, $ p(\theta;\cdot,\cdot)\in\text{MK}\;(E,\mathcal{E}) $, where $ \text{MK}\; (E,\mathcal{E}) $ is the set of all Markov kernels on $ (E,\mathcal{E}) $.
			\item[(2)] For every fixed $ A\in\mathcal{E} $, $ p(\cdot;\cdot,A) $ on $ \mathcal{Y}\times\mathcal{X} $ is measurable.
		\end{itemize}
		Then $ p $ is said to be a random Markov kernel. The set of all random Markov kernels on $ (E,\mathcal{E}) $ is noted as $ \text{RMK}(E,\mathcal{E}) $.
		
		In particular, if $ E $ is a finite set, then any random Markov kernel $ p(\theta;i,A) $ is given by a random transition matrix $ P(\theta):=(p (\theta;i,j),i,j\in E ), $ where $ p (\theta;i,j):=p (\theta;i,\{j\}) $, and
		$ p (\theta;i,A)=\sum_{j\in A} p (\theta;i,j). $
	\end{definition}
	\begin{definition}[Markov chains in random environments]\label{defMCRE}
		Let $ \vec{X}=\{X_{n},n\in T_{1}\} $ and $ \vec{\xi}=\{\xi_{n},n\in T_{2}\} $ be two random sequences defined on the probability space $ (\Omega,\mathcal{F},\mathbf{P}) $ taking values in the finite sets $ E $ and $ Y $, respectively. $  p(\cdot;\cdot,\cdot) $ is a random Markov kernel. If for any $ n\ge 2 $ and any $ i_{0},\ldots,i_{n}\in E $, we have
		$$ \begin{aligned}
			\mathbf{P}(X_{n}=i_{n}| X_{0}=i_{0},\cdots,X_{n-1}&=i_{n-1},\vec{\xi}\;)\\&=p(\xi_{n-1};i_{n-1},i_{n}).
		\end{aligned} $$
		Then we say that $ (\vec{X},\vec{\xi}) $ is a Markov chain in a positive time-dependent random environment, which is abbreviated as Markov chain in a random environment in this paper, and is denoted as MCRE.
	\end{definition}

	If $ \vec{\xi}=\{\xi_{n},n\in T_{2}\} $ is not stochastic, then the MCRE degenerates to a classical Markov chain. The MCRE is the Markov chain whose transition probability is influenced by a random environment.
	
	We are also interested in the rate that the initial distribution over state space converge to the stationary distribution. In Markov chains theory, the mixing time is used to denote the time required for a certain probability distribution to converge to the stationary distribution.
	
	\begin{definition}[mixing time]
		Let $ \vec{X}=\{X_{n},n\in T\} $ be a time-homogeneous Markov chain on a finite state space $ E $ with transition matrix $ P $ and stationary distribution $ \pi $. The mixing time is defined by
		$$ t_{mix}(\epsilon):=\min\{t:d(t)\leq\epsilon\}, $$
		where
		$$ d(t):=\max_{i\in E}\|P^{t}(i,\cdot)-\pi\|_{TV}. $$
	\end{definition}
	\begin{proposition}\label{prop3}
		Let $ P $ be the transition matrix of a Markov chain with state space $ E $ and let  $ \mu $ and $ \nu $ be any two distributions on $ E $. Then
		$$ \|\mu P-\nu P\|_{TV}\leq \|\mu-\nu\|_{TV}. $$
		This in particular shows that $ \|\mu P^{t+1}-\pi\|_{TV}\leq \|\mu P^{t}-\pi\|_{TV} $, that is, advancing the chain can only move it closer to stationarity.
	\end{proposition}
	\section{Connection between GNNs and Markov chains}\label{3}
	In section we establish a connection between GNNs and Markov chains on the graph. This is the mathematical framework for our study of the over-smoothing problem.
	\subsection{Message passing framework}\label{3.1}
	In this subsection we connect MPNN with a Markov chain on the graph and divide it into two categories of models.
	
	Recalling the message passing framework equation (1), since the node features at the $ l $th layer are only obtained from the node features at the $ l-1 $th layer, independent of the previous $ l-2 $ layers, the message passing process can be described as a Markov chain on the graph. 
	
	We take the node set $ \mathcal{V} $ as the state space and construct the family of transition matrices $ \{P^{(l)}\} $ according to functions $ \text{MSG}^{(l)} $ and $ \text{UP}^{(l)} $. Consider the node features $ H^{(l)} $ as the distribution on the node set $ \mathcal{V} $. Then the message passing process at $l$th layer is a one-step transition process of the distribution $ H^{(l-1)} $ according to the one-step transtion matrix $ P^{(l)} $. In this way, we establish the relationship between MPNN and a Markov Chain $ \vec{V} $ with state space of $ \mathcal{V} $ and initial distribution $ H^{(0)} $, transferring according to the family of transition matrices $ \left\{P^{(l)}\right\} $. Then we can use Markov chains theory to study the message passing model. 
	
	If the message passing operator is consistent at each layer, i.e., for all $ l\ge 1 $, $ \text{MSG}^{(l)} $ and $ \text{UP}^{(l)} $ are the same, then the transition matrices $ P^{(l)},\ \forall l\ge 1 $ are the same and we can connect it with a time-homogeneous Markov chain on the graph. Otherwise, the message passing operator is inconsistent, and we can connect it with a time-inhomogeneous Markov chain. We classify GNNs into two categories, \emph{operator-consistent GNN} and \emph{operator-inconsistent GNN}, based on whether the message passing operators of each layer of the model are consistent. Specifically, we use GCN and GAT as representatives of operator-consistent GNN and operator-inconsistent GNN, respectively, which are discussed in Sections \ref{3.2} and \ref{3.3}.
	\subsection{Graph convolution network}\label{3.2}
	In this subsection, we discuss the relationship between GCN and a simple random walk on the graph. 
	
	Considering $ \vec{V} $ is a simple random walk on $ \tilde{\mathcal{G}} $ with transition matrix $ \tilde{P}_\text{rw}=\tilde{D}^{-1}\tilde{A}. $ The graph convolution operator can be written as $ P_{\text{GCN}}=\tilde{D}^{-\frac{1}{2}}\tilde{P}_\text{rw}^{\text{T}}\tilde{D}^{\frac{1}{2}}. $ Thus the message passing \footnote{Similar to \cite{li2018deeper}, in this paper's discussion of graph neural networks, we focus on message passing of GNNs and omit the nonlinear activation function $ \sigma $ between layers in GNNs. In this paper, we call this expression in the form of equation (5) the message passing of the model.} of the $ l \; $th layer of GCN is
	$$ H^{(l)}=P_{\text{GCN}}H^{(l-1)}=\tilde{D}^{-\frac{1}{2}}\left(\tilde{P}_\text{rw}^{\text{T}}\right)^{l}\tilde{D}^{\frac{1}{2}}H^{(0)}. \eqno{(5)}$$
	Both sides multiply left $ \tilde{D}^{\frac{1}{2}} $ at the same time $ \tilde{D}^{\frac{1}{2}}H^{(l)}=\left(\tilde{P}_\text{rw}^{\text{T}}\right)^{l}\tilde{D}^{\frac{1}{2}}H^{(0)}. $ Let $ X^{(l)}=(\tilde{D}^{\frac{1}{2}}H^{(l)})^{\text{T}}\in\mathbb{R}^{F\times N} $, then for both sides transposed simultaneously we have
	$$ X^{(l)}=X^{(l-1)}\tilde{P}_\text{rw}. \eqno{(6)} $$
	This is exactly the simple random walk $ \vec{V}_{\text{rw}} $ with initial distribution $ X^{(0)} $ on the graph $ \tilde{\mathcal{G}} $.
	
	Combining the above discussion, we connect the most basic model in GNN, vanilla GCN, with a simple random walk on $ \tilde{\mathcal{G}} $ which is the simplest Markov chain on the graph. The graph convolution models \cite{2016Semi,atwood2016diffusion,simonovsky2017dynamic,pham2017column} are constructed by designing graph convolution operators and then stacking the graph convolution layers. We can follow the above discussion to view the graph convolution operator as one-step transition matrix and connect graph convolution models with time-homogeneous Markov chains on graph $ \mathcal{G} $. Accordingly, we can analyze the graph convolution model with the help of findings in the time-homogeneous Markov chain.
	
	\subsection{Graph attention network}\label{3.3}
	In this subsection, we discuss the relationship between GAT and a time-inhomogeneous Markov chain on the graph. 
	
	Recalling the definition of GAT, since $ p_{\text{att}}^{(l)}(u,v)\ge0 $ and $ \sum_{v=1}^{N}p_{\text{att}}^{(l)}(u,v)=1 $, $ P_{\text{att}}^{(l)} $ is a one-step stochastic matrix of a random walk on the graph. Moreover, since $ P_{\text{att}}^{(l)},\ l=1,2,\ldots $ is not the same, the forward propagation process of node representations in GAT is a time-inhomogeneous random walk on a graph, denoted as $ \vec{V}_{\text{att}} $. It has the state space $ \mathcal{V} $ and the family of stochastic matrices $ \left\{P_{\text{att}}^{(1)},P_{\text{att}}^{(2)},\ldots,P_{\text{att}}^{(l)},\ldots\right\} $. The inconsistency of the nodes message passing at each layer in GAT causes the time-inhomogeneousness of the corresponding chain $ \vec{V}_{\text{att}} $, which is an important difference between GAT and GNNs with consistent message passing such as GCN.
	
	
	
	\subsection{DropEdge method}\label{3.5}
	In this subsection, we connect DropEdge method with a random environment. Specifically, we discuss the relationship between DropEdge+GCN\footnote{DropEdge+GCN denotes the GCN model using DropEdge method} and a random walk in a random environment on the graph.
	
	Consider the stochastic process $$ \vec{\xi}=(\xi_{1}, \xi_{2},\ldots,\xi_{l},\ldots):=(\Theta^{(1)},\Theta^{(2)},\ldots,\Theta^{(l)},\ldots) $$
	with the set of time parameters $ T_{2}=\mathbb{Z}^{+} $, and taking values in $ \vec{\Xi}:=\Xi_{1}\times\Xi_{2}\times\cdots\times\Xi_{l}\times\cdots $, where $ \Xi_{i}\subset\{0,1\}^{N\times N},\forall i\ge 1 $. $ \Theta^{(l)},\ l\ge 1 $ are the random adjacency matrices that are independent and have the same distribution. If $ (u,v)\in\tilde{\mathcal{E}} $, its elements $ \theta^{(l)}(u,v) $ satisfy
	$$ \mathbf{P}(\theta^{(l)}(u,v))=\begin{cases}
		1-\dfrac{1}{|\tilde{\mathcal{E}}|}&\theta^{(l)}(u,v)=1\\
		\dfrac{1}{|\tilde{\mathcal{E}}|}&\theta^{(l)}(u,v)=0
	\end{cases} $$
	is a Bernoulli random variable. It indicates that each edge $ (u,v) $ is dropped with a uniform probability $ \frac{1}{|\tilde{\mathcal{E}}|} $. We use the random environment $ \vec{\xi}=(\Theta^{(1)},\Theta^{(2)},\ldots,\Theta^{(l)},\ldots) $ to model $ \left(A^{(1)}_{\text{drop}},A^{(2)}_{\text{drop}},\ldots,A^{(l)}_{\text{drop}},\ldots\right) $.
	
	Next we connect the DropEdge+GCN with a simple random walk on the graph in the random environment. The message passing of DropEdge+GCN at the $ l $th layer is
 $ H^{(l)} = \left(D_{\text{drop}}^{(l)}\right)^{-1}A^{(l)}_{\text{drop}}H^{(l-1)}, $ where the degree matrix after DropEdge is $ D_{\text{drop}}^{(l)}:=diag(\zeta^{(l)}_{1},\ldots,\zeta^{(l)}_{N}) $, the degree of node $ u $ 
	$$ \zeta^{(l)}_{u}:=\sum_{v\in\mathcal{N}(u)}\theta^{(l)}(u,v) \eqno{(7)}$$
	is a random variable. Since $ \theta^{(l)}(u,v) $ is a Bernoulli random variable, the random variable $ \zeta^{(l)}_{u} $ follows a binomial distribution with parameters $ |\mathcal{N}(u)|=\deg(u) $ and $ 1-\frac{1}{|\mathcal{E}|} $, i.e. $ \zeta^{(l)}_{u}\sim B(\deg(u),1-\frac{1}{|\mathcal{E}|}). $ Let $ \vec{V}=(V_{0},V_{1},\ldots,V_{l},\ldots) $ be the original chain. $ V_{i},i=0,1,\ldots $ is random variables taking values in $ \mathcal{V} $. Consider the random transition matrix
	$ P(\Theta^{(l)}):=\tilde{D}_{\Theta^{(l)}}^{-1} \Theta^{(l)}, $ where $ \tilde{D}_{\Theta^{(l)}}=D_{\text{drop}}^{(l)} $. Since $ \forall l\ge 2 $ and $ \forall v_{0},v_{1},\ldots,v_{l}\in \mathcal{V} $,
	$$ \begin{aligned}
		\mathbf{P}(V_{l}=v_{l}|V_{0}&=v_{0},\ldots,V_{l-1}=v_{l-1},\vec{\xi})\\&=p(\Theta^{(l)};v_{l-1},v_{l})=\frac{\theta^{(l)}(v_{l-1},v_{l})}{\zeta^{(l)}_{v_{l-1}}},
	\end{aligned} $$
	according to Definition \ref{defMCRE}, $ (\vec{V},\vec{\xi}) $ is a MCRE.
	
	DropEdge applied to different GNN models means MCRE with the same random environment but different original chains. Although the specific details are not same, the modeling methods are the same. Next, in Section \ref{4.2}, we discuss in depth the effectiveness of DropEdge method to alleviate the over-smoothing problem.

\section{Markov analysis of the over-smoothing problem}\label{4}
	In this section, we comprehensively analyze the over-smoothing problem in general GNN from the Markov chain perspective. First, in \ref{4.1} we reveal the cause of over-smoothing. Based on this, in \ref{4.2} we show that although the previously proposed methods can alleviate over-smoothing, these methods cannot entirely avoid the over-smoothing problem. Next, in \ref{4.3} we explore what types of GNNs can avoid the over-smoothing problem. Finally, in \ref{4.4} we give a solution to the over-smoothing problem in the Markovian sense.

	
	\subsection{Cause of over-smoothing}\label{4.1}
	In this subsection, we state the cause of over-smoothing problem. In \ref{3.1} we show that the message passing framework can be related to a Markov Chain on the graph. Meanwhile, the feature vectors of the nodes can be viewed as a discrete probability distribution over the node set. The forward propagation process of node representation is the transfer process of the distribution on the node set $ \mathcal{V} $. As GNN propagates forward, the representation distribution converges to the stationary distribution. This causes the over-smoothing problem in GNN. Next, we take vanilla GCN as an example to specifically analyze why the node representations are over-smoothing.
	
	For a simple random walk on the graph $ \tilde{\mathcal{G}} $, for all $ v\in\tilde{\mathcal{V}} $
	$$ \sum_{u\in\tilde{\mathcal{V}}}\deg(u)\tilde{p}_{\text{rw}}(u,v)=\sum_{(u,v)\in\tilde{\mathcal{E}}}\frac{\deg(u)}{\deg(u)}=\deg(v), $$
	where $ \tilde{p}_{\text{rw}}(u,v)=\frac{1}{\deg(u)} $ if $ (u,v)\in\tilde{\mathcal{E}} $ otherwise $ \tilde{p}_{\text{rw}}(u,v)=0 $. To get a probability, we simply normalize by $ \sum_{v\in\tilde{\mathcal{V}}}\deg(v)=2|\tilde{\mathcal{E}}| $. Then the probability measure
	$$ \pi(u)=\frac{\deg(u)}{2|\tilde{\mathcal{E}}|}\quad\forall u\in\tilde{\mathcal{V}} $$
	is always a stationary for the walk. Write in matrix form
	$ \pi=\pi \tilde{P}_{\text{rw}} $.
	Recall in \ref{3.2} we connect the vanilla GCN with a simple random walk on the graph. Noting $ X^{(l)}=\left(\tilde{D}^{\frac{1}{2}}H^{(l)}\right)^{\text{T}} $, the message passing of the $ l \; $th GCN hidden layer is
	$$ X^{(l)}=X^{(l-1)}\tilde{P}_{\text{rw}}. $$
	Since simple random walks on connected graphs are irreducible and aperiodic Markov chains,
	$$ X^{(l)}(k,\cdot)\rightarrow\pi,\quad l\rightarrow\infty,  $$
	where $ X^{(l)}(k,\cdot),\ k=1,2,\ldots,F $ is the distribution over the node set $ \tilde{\mathcal{V}} $ consisting of the $ k $th component of each node feature, $ \pi $ is the unique stationary distribution of $ \tilde{P}_{\text{rw}} $. Thus for all $ u\in \tilde{\mathcal{V}} $
	$$ x^{(l)}(1,u)=x^{(l)}(2,u)=\cdots=x^{(l)}(F,u)=\pi(u), $$
	where $ \pi(u) $ is the $ u $th component of $ \pi $. 
	
	From the above discussion, the over-smoothing problem is due to the fact that the node representation distribution converges to be stationary distribution, resulting in the consistency of the each node representation. 
	
	\subsection{Lazy walk analysis of previous methods}\label{4.2}
	In this subsection, we first uniformly connect the methods that alleviate the over-smoothing problem including the residual connections method \cite{2016Semi,chiang2019cluster}, PPNP \cite{klicpera2018predict}, and the DropEdge method \cite{rong2019dropedge,huang2020tackling} with lazy walks on the graph. Next, we prove the effectiveness of previous methods to alleviate the over-smoothing. Finally, we show that these methods cannot entirely avoid the over-smoothing problem in GNN.
	
	\noindent
	\textbf{Residual Connections Method.} If we omit the nonlinear activation function in the forward propagation process of the graph neural network and focus only on the node message passing process, then the models of equation (2) and (3) are the same. The operator $ P_{\text{GCN}}+I $ is normalized
	$$ P_{\text{res}}:=\frac{1}{2}P_{\text{GCN}}+\frac{1}{2}I. $$
	Then the message passing for the residual connections method is
	$$ H^{(l)}=\left(\frac{1}{2}P_{\text{GCN}}+\frac{1}{2}I\right)H^{(l-1)}=P_{\text{res}}H^{(l-1)}. $$
	Since $ P_{\text{res}} $ is the transition matrix of a lazy walk with parameter $ \gamma=\frac{1}{2} $, residual connections method can be related to a lazy walk on the graph.
	
	\noindent
	\textbf{PPNP.} We consider the lazy walk with the parameter $ \alpha $ on $ \tilde{\mathcal{G}} $ with transition matrix
	$$ P_{\text{lazy}}=(1-\alpha)\tilde{D}^{-1}\tilde{A}+\alpha I. \eqno{(8)}$$
	This is formally similar to the PPNP messaging (equation (4)). Intuitively, equation (8) indicates that the original random walk has a $ 1-\alpha $ probability of continuing to walk, and a $ \alpha $ probability of staying in place. This is exactly the idea of personalized PageRank.
	
	Combining the above discussion, we connect PPNP with a lazy walk on $ \tilde{\mathcal{G}} $ with parameter $ \alpha $. As can be seen from the message passing expression (8), PPNP is a more general method than the residual connections method. In particular, if $ \alpha= \frac{1}{2} $, PPNP degenerates to the residual connections method.
	
	\noindent
	\textbf{DropEdge.} Unlike residual connections method and PPNP, DropEdge method explicitly have nothing to do with the form of lazy walk. However, we will prove that message passing of the DropEdge+GCN model is a lazy walk on the graph. In \ref{3.5} we connect DropEdge+GCN with a MCRE $ (\vec{V},\vec{\xi}\;) $ on the graph. The following theorem illustrates that the original chain $ \vec{V} $ is a lazy walk on the graph.
	
	\begin{theorem}\label{thm4.2}
		Let $ (\vec{V},\vec{\xi}) $ be the MCRE that describes the DropEdge+GCN model in \ref{3.5}. $ \vec{\xi}=(\Theta^{(1)},\Theta^{(2)},\ldots,\Theta^{(l)},\ldots) $ is a random environment with independent identical distribution. Then the original chain $ \vec{V} $ is a time-homogeneous Markov chain with a transition matrix
		$$ P_{\text{drop}}:=(I-\Gamma)\tilde{D}^{-1}\tilde{A}+\Gamma, \eqno{(9)}$$
		where $ \Gamma:=diag\left(\frac{1}{|\mathcal{E}|^{\deg(1)}},\frac{1}{|\mathcal{E}|^{\deg(2)}},\ldots,\frac{1}{|\mathcal{E}|^{\deg(N)}}\right) $
		is a diagonal matrix.
	\end{theorem}

	The conclusion of Theorem \ref{thm4.2} tells us the relationship between DropEdge+GCN and a lazy walk on the graph. Equation (9) is intuitively equivalent to that the message of node $ u $ passes with the probability $ 1-\frac{1}{|\mathcal{E}|^{\deg(u)}} $, and stays in node $ u $ with probability $ \frac{1 }{|\mathcal{E}|^{\deg(u)}} $.
	
	In fact, equation (9) is a more generalized form of lazy walk on the graph. In particular, if $ \mathcal{G} $ is a regular graph, i.e., a graph with the same degree of each node. Then the matrix $ \Gamma:=diag(\frac{1}{|\mathcal{E}|^{\deg(1)}},\frac{1}{|\mathcal{E}|^{\deg(2)}},\ldots,\frac{1}{|\mathcal{E}|^{\deg(N)}}) $ degenerates to the constant $ \frac{1}{|\mathcal{E}|^{\deg(u)}} $. Then the message passing of the DropEdge+GCN model is a lazy walk with the parameter $ \frac{1}{|\mathcal{E}|^{\deg(u)}} $ on the graph.
	
	We have already uniformly connected previously proposed methods with lazy walk on graphs. Next we will use the mixing time theory as a mathematical tool to analyze the property of the lazy walk on the graph. This is used to illustrate the effectiveness of these methods to alleviate the over-smoothing problem.
	
	The following Theorem \ref{thm4.2.1} illustrates that starting from the any initial distribution, the lazy walk on the graph moves to the stationary distribution more slowly than the simple random walk.
	\begin{theorem}\label{thm4.2.1}
		Let $ P_{\text{rw}}:=D^{-1}A $ be the transition matrix of a simple random walk $ \vec{V}_{\text{rw}} $ on the graph $ \mathcal{G}=(\mathcal{V},\mathcal{E}) $, $ P_{\text{lazy}}:=(1-\gamma)D^{-1}A+\gamma I $ be the transition matrix of the lazy walk $ \vec{V}_{\text{lazy}} $ on the graph $ \mathcal{G}=(\mathcal{V},\mathcal{E}) $. If $ \vec{V}_{\text{rw}} $ and $ \vec{V}_{\text{lazy}} $ move from any distribution on $ \mathcal{V} $, then there is the following conclusion
		\begin{itemize}
			\item[(1)]$ \vec{V} $ and $ \vec{V}_{\text{lazy}} $ have the same stationary distribution $ \pi $, where
			$ \pi(u)=\frac{\deg(u)}{2|\mathcal{E}|},\quad\forall u\in\mathcal{V}. $
			\item[(2)] For all $ l>0 $,
			$$ \max_{u\in\mathcal{V}}\|P^{l}_{\text{lazy}}(u,\cdot)-\pi\|_{TV}\ge \max_{u\in\mathcal{V}}\|P_{\text{rw}}^{l}(u,\cdot)-\pi\|_{TV}, $$
			where $ \|\cdot\|_{TV} $ is the total variation distance.
		\end{itemize}
	\end{theorem}
	Notice the definition of mixing time in Section \ref{2}
	$$ t_{mix}(\epsilon):=\min\{t:d(t)\leq\epsilon\},\ d(t):=\max_{u\in\mathcal{V}}\|P^{l}(u,\cdot)-\pi\|_{TV}. $$
	Theorem \ref{thm4.2.1} shows that the mixing time for a lazy walk on the graph moving to the stationary distribution is longer than that for a simple random walk. 
	
	Back to the over-smoothing problem in GCN, in \ref{4.1}, we attribute over-smoothing to the convergence of the probability distribution to the stationary distribution. Theorem \ref{thm4.2.1} shows that the residual connections method, PPNP and DropEdge method which can be related to lazy walks on the graph can indeed alleviate the over-smoothing problem.
	
	Finally, we discuss the problem that can these methods entirely avoid the over-smoothing. The following Theorem \ref{thm4.2.2} shows that for a lazy walk on the graph, the probability distribution on the node set still converges to the stationary distribution, and the rate of convergence is exponential as with a simple random walk on the graph.
	\begin{theorem}\label{thm4.2.2}
		Let $ L:=I-D^{-\frac{1}{2}}AD^{-\frac{1}{2}} $ be the normalized Laplacian matrix of the graph $ \mathcal{G} $, $ \lambda_{1}\leq \lambda_{2}\leq\cdots\leq\lambda_{N} $ be the eigenvalues of $ L $, $ \phi_{1},\phi_{2},\cdots,\phi_{N} $ are the corresponding eigenvectors of the corresponding eigenvalues. Then for any initial distribution $ \mu:\mathcal{V}\rightarrow\mathbb{R} $, $ l>0 $,
		$$ \mu P_{\text{rw}}^{l}=\pi+\sum_{k=2}^{N}(1-\lambda_{k})^{l}a_{k}\phi_{k}D^{\frac{1}{2}}, $$
		$$ \mu P_{\text{lazy}}^{l}=\pi+\sum_{k=2}^{N}(1-(1-\gamma)\lambda_{k})^{l}a_{k}\phi_{k}D^{\frac{1}{2}}, $$
		where $ a_{k},k=1,2,\ldots,N $ are the coordinates of the vector $ \mu D^{-\frac{1}{2}} $ on the 
		basis $ (\phi_{1},\phi_{2},\cdots,\phi_{N}) $, i.e. $ \mu D^{-\frac{1}{2}}=\sum_{k=1}^{N}a_ {k}\phi_{k} $. Further, by the Frobenius-Perron Theorem, we have $ 0=\lambda_{1}\leq \lambda_{2}\leq\cdots\leq\lambda_{N}\leq 2 $, so $ \mu P^{l} $ and $ \mu P^{l}_{\text{lazy}} $ both converge to $ \pi $ with $ l $ exponentially.
	\end{theorem}
	
	In this subsection, we use Theorem \ref{thm4.2.1} to show the effectiveness of the residual connections method, PPNP and DropEdge method to alleviate the over-smoothing problem. However, it is further shown by Theorem \ref{thm4.2.2} that these methods can neither make GCN avoid over-smoothing nor can they make GCN avoid over-smoothing at the exponential rate. Then there is a problem that what types of GNNs can avoid the over-smoothing problem. We will discuss in detail in the next subsection.
	\subsection{Conclusion of the general GNN model}\label{4.3}
	In this subsection, we give the conclusions of the over-smoothing problem in two types of GNNs in the Markovian sense. In \ref{4.3.1}, we show that operator-consistent GNN cannot avoid over-smoothing at an exponential rate. In \ref{4.3.2}, we state that operator-inconsistent GNN is not always over-smoothing.
	
	\subsubsection{Operator-consistent GNN models}\label{4.3.1}
	We discuss following two core issues about operator-consistent GNN.
	\begin{itemize}
		\item Can operator-consistent GNN models avoid over-smoothing?
		\item If over-smoothing cannot be avoided, can operator-consistent GNN models avoid over-smoothing at the exponential rate?
	\end{itemize}
	The answer to both questions is no. The following Theorem \ref{thm4.3.1} answers the first question by showing that the stationary distribution must exist for the transition matrix of time-homogeneous random walk on the graph. 
	\begin{theorem}\label{thm4.3.1}
		Let $ P $ be a transition matrix of a time-homogeneous random walk on the connected, non-bipartite graph $ \mathcal{G} $, Then there exists a unique probability distribution $ \pi $ over $ \mathcal{V} $ that satisfies $ \pi=\pi P. $
	\end{theorem}
	
		
		Theorem \ref{thm4.3.1} shows that there must be a stationary distribution for the message passing operator on the graph. Then distribution on $ \mathcal{V}$ will converge to stationary distribution $ \pi $. For an operator-consistent GNN, the node representations will be over-smoothing as the GNN propagates forward. Combining the above discussion, the operator-consistent GNN cannot avoid over-smoothing.
		
		The following Theorem \ref{thm4.3.2} answers the second question by showing that the time-homogeneous random walk on the graph will converge at the exponential rate to its stationary distributions $ \pi $.
		\begin{theorem}\label{thm4.3.2}
			Under the condition of Theorem \ref{thm4.3.1}, there exist constants $ \alpha\in(0,1) $ and $ C>0 $ such that
			$$ \max_{u\in\mathcal{V}}\|P^{l}(u,\cdot)-\pi\|_{TV}\leq C\alpha^{l}, $$
			where $ P(u,\cdot) $ denotes the row vector consisting of the $ i $th row of the transition matrix $ P $.
		\end{theorem}
		
		Theorem \ref{thm4.3.2} shows that operator-consistent GNN will be over-smoothing at the exponential rate. Thus as long as the message passing operators of each layer are consistent, the GNN model cannot avoid over-smoothing at the exponential rate.
		
		To summarize the above discussion, the operator-consistent GNN model can neither avoid over-smoothing nor can it avoid over-smoothing at the exponential rate in the Markovian sense. The over-smoothing problem can only be alleviated but not be avoided.
		
		\subsubsection{Operator-inconsistent GNN models}\label{4.3.2}
		We take GAT which is the most typical operator-inconsistent GNN as an example. We first show that the conclusion that the GAT will be over-smoothing cannot be proven.
		
		The following theorem gives property of the family of stochastic matrices $ \left\{P_{\text{att}}^{(1)},P_{\text{att}}^{(2)},\ldots,P_{\text{att}}^{(l)},\ldots\right\} $, shows the existence of stationary distribution of each graph attention matrix, and gives the explicit expression of stationary distribution.
		\begin{theorem}\label{thm4.3.3}
			There exists a unique probability distribution $ \pi^{(l)} $ on $ \mathcal{V} $ satisfies
			$$ \pi^{(l)}=\pi^{(l)} P_{\text{att}}^{(l)},\quad l=1,2,\ldots, $$
			where $ \pi^{(l)}(u)=\frac{\deg^{(l)}(u)}{\sum_{k\in\mathcal{V}}\deg^{(l)}(k)} $, $ \deg^{(l)}(u)=\sum_{z\in\mathcal{N}(u)}\exp(\phi^{(l)}(h_{u}^{(l-1)},h_{z}^{(l-1)})) $.
		\end{theorem}
		Previously, \cite{wang2019improving} discussed the over-smoothing problem in GAT and concluded that the GAT would over-smooth. Same as our work, they viewed the $ P_{\text{att}}^{(l)} $ at each layer as stochastic matrix of a random walk on the graph. However, they ignore the fact that the complete forward propagation process of GAT is essentially a time-inhomogeneous random walk on the graph. The core theorem stating that the GAT will over-smooth in their work is flawed. In its proof, the stationary distribution $ \pi^{(l)} $ of the graph attention matrix $ P_{\text{att}}^{(l)} $ for each layer is consistent, i.e. $$ \pi^{(1)}=\pi^{(2)}=\cdots=\pi^{(l)}=\cdots. $$ However, since each layer $ \phi^{(l)} $ is different, by Theorem~\ref{thm4.3.3},
		$$ \pi^{(1)}\neq\pi^{(2)}\neq\cdots\neq\pi^{(l)}\neq\cdots. $$
		The conclusion that the GAT will be over-smoothing cannot be proven. Then we show that the GAT is not always over-smoothing.
		
		Recalling Lemma \ref{lem4}, for time-inhomogeneous random walk $ \vec{V}_{\text{att}} $, its family of stochastic matrices $ \{P_{\text{att}}^{(l)}\} $ satisfies the condition (1) (Theorem \ref{thm4.3.3}). However, the series of positive terms $ \sum_{l}\|\pi^{(l)}-\pi^{(l+1)}\| $ is possible to be divergent and the condition $ (2) $ of Lemma \ref{lem4} can not be guaranteed. Moreover, according to the definition of $ \{P_{\text{att}}^{(l)}\} $, neither the Isaacson-Madsen condition nor the Dobrushin condition can be guaranteed. So the time-inhomogeneous chain $ \vec{V}_{\text{att}} $ does not always have a limiting distribution. This indicates that GAT is not always over-smoothing.
		
		Since the Lemma \ref{lem4} holds for all time-inhomogeneous Markov chains, the analysis on GAT can be generalized to the general operator-inconsistent model. To summarize the discussion, we conclude that operator-inconsistent GNNs are not always over-smoothing.
		
		\subsection{Sufficient condition to avoid over-smoothing}\label{4.4}
		In this subsection, we propose and prove a necessary condition for the existence of stationary distribution for a time-inhomogeneous Markov chain. Then we apply this theoretical result to operator-inconsistent GNN and propose a sufficient condition to ensure that the model can avoid over-smoothing.
		
		In the study of Markov chains, researchers usually focus on the sufficient conditions for the existence of the limiting distribution. And the case when the limiting distribution does not exist has rarely been studied. We study the necessary conditions for the existence of the limit distribution in order to obtain sufficient conditions for its nonexistence.
		
		The following theorem gives a necessary condition for the existence of the stationary distribution of the time-inhomogeneous Markov chain. Although other necessary conditions exist, Theorem \ref{thm5.3} is one of the most intuitive and simplest in form.
		
		\begin{theorem}\label{thm5.3}
			Let $ \vec{X}=\{X_{n},n\in T\} $ be a time-inhomogeneous Markov chain on a finite state space $ E $, and write its $ n $th-step transition matrix as $ P^{(n)} $, satisfying that, $ P^{(n)} $ is irreducible and aperiodic, there exists a unique stationary distribution $ \pi^{(n)} $ as the time-homogeneous transition matrix, and
			$  C(P^{(n)})<1 $.
			Let the initial distribution be $ \mu_{0} $ and the distribution of the chain $ \vec{X} $ at step $ n $ be $ \mu_{n}:= \mu_{n-1}P^{(n)} $. Then the necessary condition for there exists a probability distribution $ \pi $ on $ E $ such that 
			$ \|\mu_{n}-\pi\|\rightarrow 0,\; n\rightarrow \infty $
			is
			$$ \|\pi^{(n)}-\pi\|\rightarrow 0,\quad n\rightarrow \infty.  $$
		\end{theorem}
		We explain Theorem \ref{thm5.3} intuitively. In the limit sense, transition of $ \mu_{n-1} $ satisfies
		$$ \lim\limits_{n\rightarrow \infty}\mu_{n-1}P^{(n)}=\lim\limits_{n\rightarrow \infty}\mu_{n}=\lim\limits_{n\rightarrow \infty}\mu_{n-1}=\pi. $$
		On the other hand, $ \forall n > 0 $, $ \pi^{(n)} $ is the unique solution of the equation $ \mu=\mu P^{(n)}. $ Thus 
		$ \lim\limits_{n\rightarrow \infty}\pi^{(n)}=\lim\limits_{n\rightarrow \infty}\mu_{n-1}=\pi. $
		
		We first consider GAT. By Theorem \ref{thm5.3}, we give the following sufficient condition for GAT to avoid over-smoothing in Markovian sense.
		\begin{table*}[t]
			\centering
			\caption{Results of GAT}
			\label{table: result of gat}
			~\\
			\resizebox{\linewidth}{!}{
				\begin{tabular}{l|c|cccccc}
					\hline
					\multirow{2}{*}{datasets}&\multirow{2}{*}{model}&\multicolumn{6}{c}{\#layers}\\
					\cline{3-8}
					&&3&4&5&6&7&8 \\
					\hline
					
					\multirow{2}{*}{Cora}&GAT&0.7773($ \pm $0.0054) &0.7602($ \pm $0.0166) &0.4821($ \pm $0.3021)&0.2774($ \pm $0.2542) &0.1672($ \pm $0.0780) &0.0958($ \pm $0.0059)\\
					&GAT-RT&0.7884($ \pm $0.0157) &0.7872($ \pm $0.0127) &0.7648($ \pm $0.0077)&0.6454($ \pm $0.2508) &0.3244($ \pm $0.2465) &0.1678($ \pm $0.0756)\\
					\multirow{2}{*}{Citeseer}&GAT&0.6643($ \pm $0.0063) &0.6541($ \pm $0.0076) &0.3472($ \pm $0.2582)&0.2474($ \pm $0.1947) &0.1768($ \pm $0.0064) &0.1902($ \pm $0.0598)\\
					&GAT-RT&0.6678($ \pm $0.0157) &0.6692($ \pm $0.0072) &0.6208($ \pm $0.0380)&0.2706($ \pm $0.1884) &0.1915($ \pm $0.0200) &0.1864($ \pm $0.0229)\\
					\multirow{2}{*}{Pubmed}&GAT&0.7616($ \pm $0.0115) &0.7534($ \pm $0.0114) &0.7653($ \pm $0.0072)&0.7468($ \pm $0.0084) &0.7468($ \pm $0.0045) &0.7076($ \pm $0.0112)\\
					&GAT-RT&0.7673($ \pm $0.0064) &0.7659($ \pm $0.0123) &0.7684($ \pm $0.0063)&0.7664($ \pm $0.0092) &0.7596($ \pm $0.0107) &0.7618($ \pm $0.0114)\\
					\multirow{2}{*}{ogbn-arxiv} &GAT&0.7117($ \pm $0.0023) &0.7144($ \pm $0.0015) &0.7061($ \pm $0.0082)&0.6396($ \pm $0.1031) &0.4307($ \pm $0.1720) &-\\
					&GAT-RT&0.7115($ \pm $0.0025) &0.7104($ \pm $0.0022) &0.7063($ \pm $0.0040)&0.6709($ \pm $0.0344) &0.5653($ \pm $0.1122) &-\\
					\hline
					
			\end{tabular}}
		\end{table*}
		\begin{corollary}[Sufficient condition to avoid over-smoothing]\label{cor5.1}
			Let $ h_{u}^{(l)} $ be the $ l\; $th hidden layer feature of node $ u\in\mathcal{V} $ in GAT, then a sufficient condition for GAT to avoid over-smoothing is that there exists $ \delta>0 $ such that for any $ l\ge2 $, satisfying
			$$ \|h_{u}^{(l-1)}-h_{u}^{(l)}\|>\delta. \eqno{(10)}$$
		\end{corollary}
		
		When Equation (10) is satisfied, the time-inhomogeneous random walk $ \vec{V}_{\text{att}} $ corresponding to GAT does not have a limiting distribution, and thus GAT avoids potential over-smoothing problems in a Markovian sense. Corollary \ref{cor5.1} has an intuitive meaning. The essence of over-smoothing is that the node representations converge with the propagation of the network. By Cauchy's convergence test, the condition exactly avoid representation $ h_{u}^{(l)} $ of the node $ u $ from converging as network deepens.
		
		Since Theorem \ref{thm5.3} generally holds for all time-inhomogeneous Markov chains, operator-inconsistent GNN such as GEN \cite{li2020deepergcn} can also obtain the sufficient conditions to avoid over-smoothing in Markovian sense. The analysis of GEN is provided in Appendix B.
	\section{Experiments}\label{5}
	In this section, we experimentally verify the correctness of our theoretical results. We rewrite the sufficient condition in the Corollary \ref{cor5.1} as a regularization term. It can be flexibly added to the training of the network. The experimental results show that our proposed condition can effectively avoid the over-smoothing problem and improve the performance of GAT. We also conduct experiments on GEN-SoftMax~\cite{li2020deepergcn} (Section \ref{sec5.4}).
 \begin{table*}[t]
	\centering
	\tiny
	\caption{GEN's performance on OGB datasets}
	\label{table:gen}
	~\\
	\resizebox{\linewidth}{!}{
		\begin{tabular}{l|c|cccc}
			\hline
			\multirow{2}{*}{datasets}&\multirow{2}{*}{model}&\multicolumn{4}{c}{\#layers}\\
			\cline{3-6}
			&&7&14&28&56 \\
			\hline
			
			\multirow{2}{*}{ogbn-arxiv}&GEN&0.7140($ \pm $0.0003)&0.7198($ \pm $0.0007) &0.7192($ \pm $ 0.0016)&- \\
			&GEN-RT&0.7181($ \pm $0.0006) &0.7204($ \pm $0.0014) &0.7220($ \pm $0.0008)&- \\
			\multirow{2}{*}{ogbg-molhiv}&GEN&0.7858($ \pm $0.0117) &0.7757($ \pm $0.0019) &0.7641($ \pm $0.0058)&0.7696($ \pm $0.0075) \\
			&GEN-RT&0.7872($ \pm $0.0083) &0.7838($ \pm $0.0024) &0.7835($ \pm $0.0010)&0.7795($ \pm $0.0027) \\
			\multirow{2}{*}{ogbg-ppa}&GEN&0.7554($ \pm $0.0073) &0.7631($ \pm $0.0065) &0.7712($ \pm $0.0071)&- \\
			&GEN-RT&0.7600($ \pm $0.0062) &0.7700($ \pm $0.0081) &0.7800($ \pm $0.0037)&- \\
			\hline
			
	\end{tabular}}
\end{table*}
\subsection{Setup}
In this subsection we briefly introduce the experimental settings. See Appendix C for more specific settings. We verify our conclusions while keeping the other hyperparameters the same\footnote{Since we do not aim to refresh state of the arts, these are not necessarily the optimal hyperparameters.} (network structure, learning rate, dropout, epoch, etc.). \\

\noindent\textbf{Variant of sufficient condition.}\quad Notice that the sufficient condition in the Corollary \ref{cor5.1} is in the form of inequality, which is not conducive to experiments. Let $ h_{u}^{(l)} $ be representation of node $ u\in\mathcal{V} $ at the hidden layer $ l $, we normalize the distance of the node representations between two adjacent layers and then let it approximate to a given hyperparameter threshold $ T\in(0,1) $, i.e., for the GNN model with $n$ layers, we obtain a regularization term
$$\text{RT} (x)= (\frac{1}{n}\sum_{l=1}^{n}(\| \text{Sigmoid}(h_{u}^{(l-1)})-\text{Sigmoid}(h_{u}^{(l)})  \|)-T)^2.$$
Since there must exist $ \delta>0 $ that satisfies $ T>\delta $, Equation (10) can be satisfied if this term is perfectly minimized. For a detailed choice of the threshold $T$, we put it in Appendix C. 

\noindent
\textbf{Datasets.}\quad In terms of datasets, we follow the datasets used in the original work of GAT \cite{velivckovic2017graph} as well as the OGB benchmark. We use four standard benchmark datasets: ogbn-arxiv~\cite{hu2020ogb}, Cora, Citeseer, and Pubmed \cite{sen2008collective},
covering the basic transductive learning tasks.

\noindent
\textbf{Implementation details.}\quad For the specific implementation, we refer to the open-source code of vanilla GAT, and models with different layers share the same settings. We use the Adam SGD optimizer \cite{kingma2014adam} with learning rate 0.01, the hidden dimension is 64, each GAT layer has 8 heads and the amount of training epoch is 500. All experiments are conducted on a single Nvidia Tesla v100.
\subsection{Results of GAT}\label{exp gat}
For record simplicity, we denote the GAT after adding the regularization term to the training as GAT-RT. To keep statistical confidence, we repeat all experiments 10 times and record the mean value and standard deviation. Results shown in Table~\ref{table: result of gat} demonstrate that almost on each dataset and number of layers, GAT-RT will obtain an improvement in the performance. Specifically, on Cora and Citeseer, GAT's performance begins to decrease drastically when layer numbers surpass 6 and 5 but GAT-RT can relieve this trend in some way. On Pubmed, vanilla GAT's performance has a gradual decline. The performance of vanilla GAT decrease 6\% when the layer number is 8. However, GAT-RT's performance keeps competitive for all layer numbers. For ogbn-arxiv, GAT-RT performs as competitive as GAT when the layer number is small but outperforms GAT by a big margin when the layer number is large. Specifically when the layer number is 6 and 7, the performance improves roughly by 3\% and 13\% respectively.

\subsection{Verification of avoiding over-smoothing}
\begin{figure*}[h]
	\centering  
	\subfigcapskip=-2pt 
	\subfigure[3-layer]{
		\includegraphics[width=0.175\linewidth]{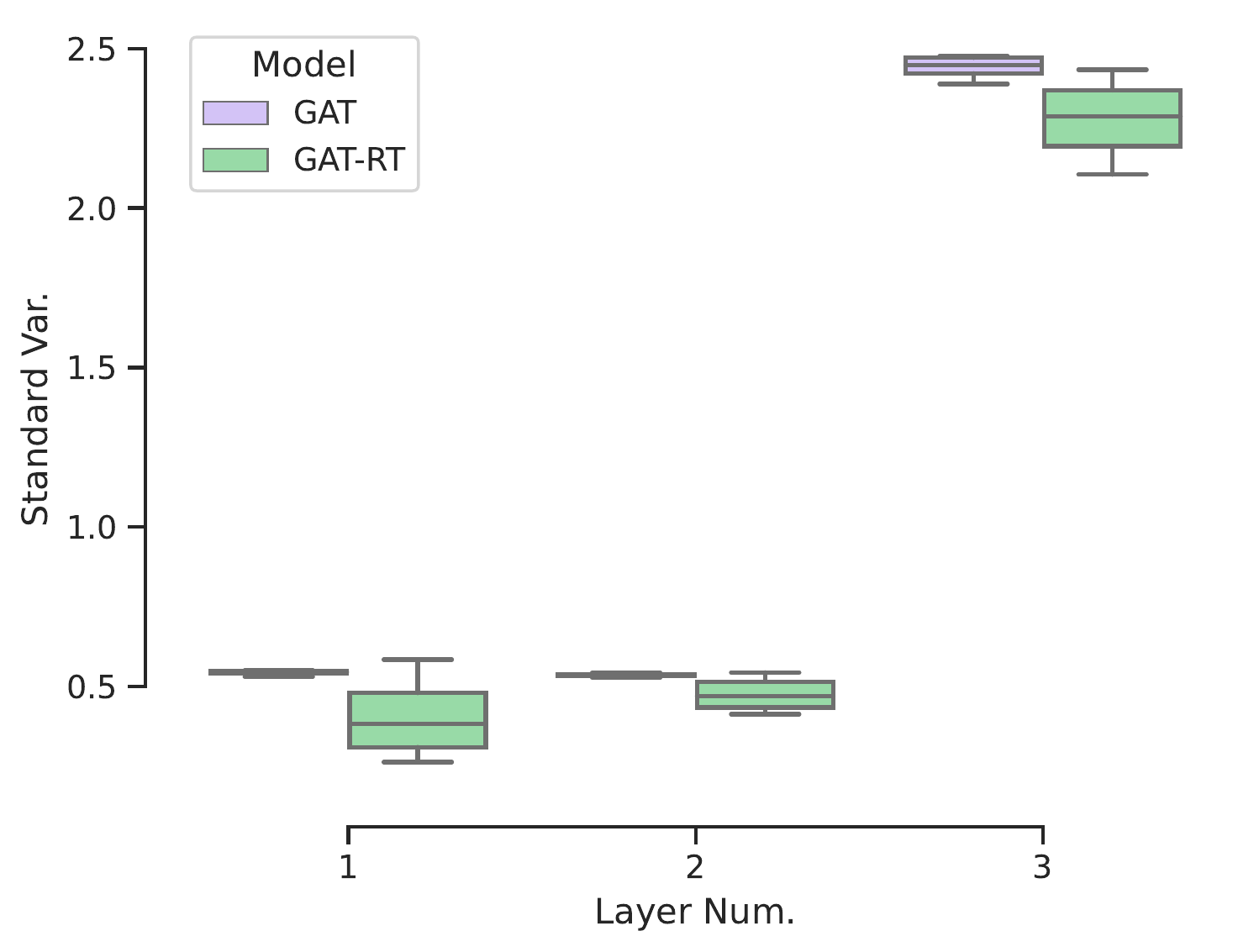}}
		\hspace{-3mm}
	\subfigure[4-layer]{
		\includegraphics[width=0.175\linewidth]{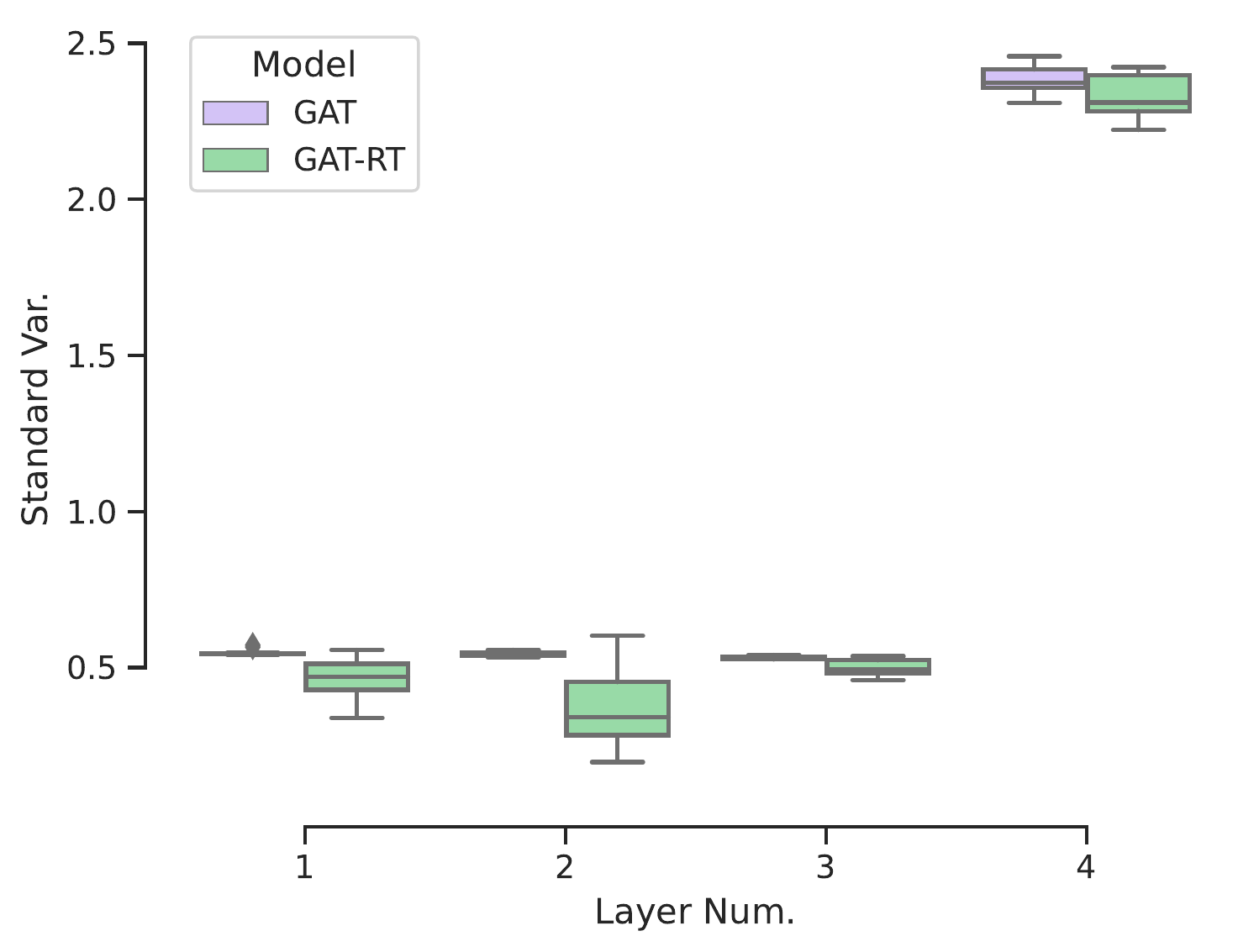}}
				\hspace{-3mm}
	\subfigure[5-layer]{
		\includegraphics[width=0.175\linewidth]{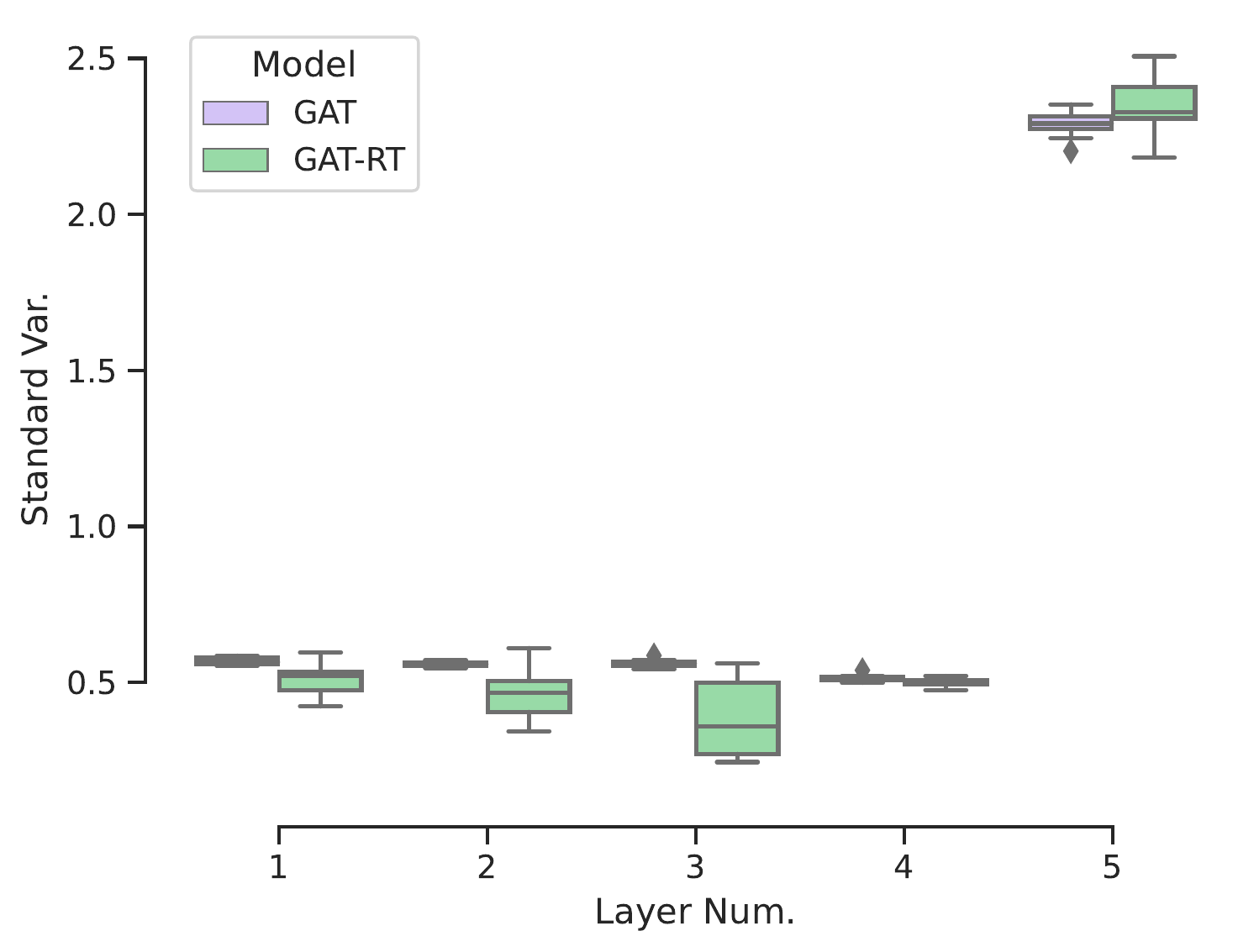}}
				\hspace{-3mm}
	\subfigure[6-layer]{
		\includegraphics[width=0.175\linewidth]{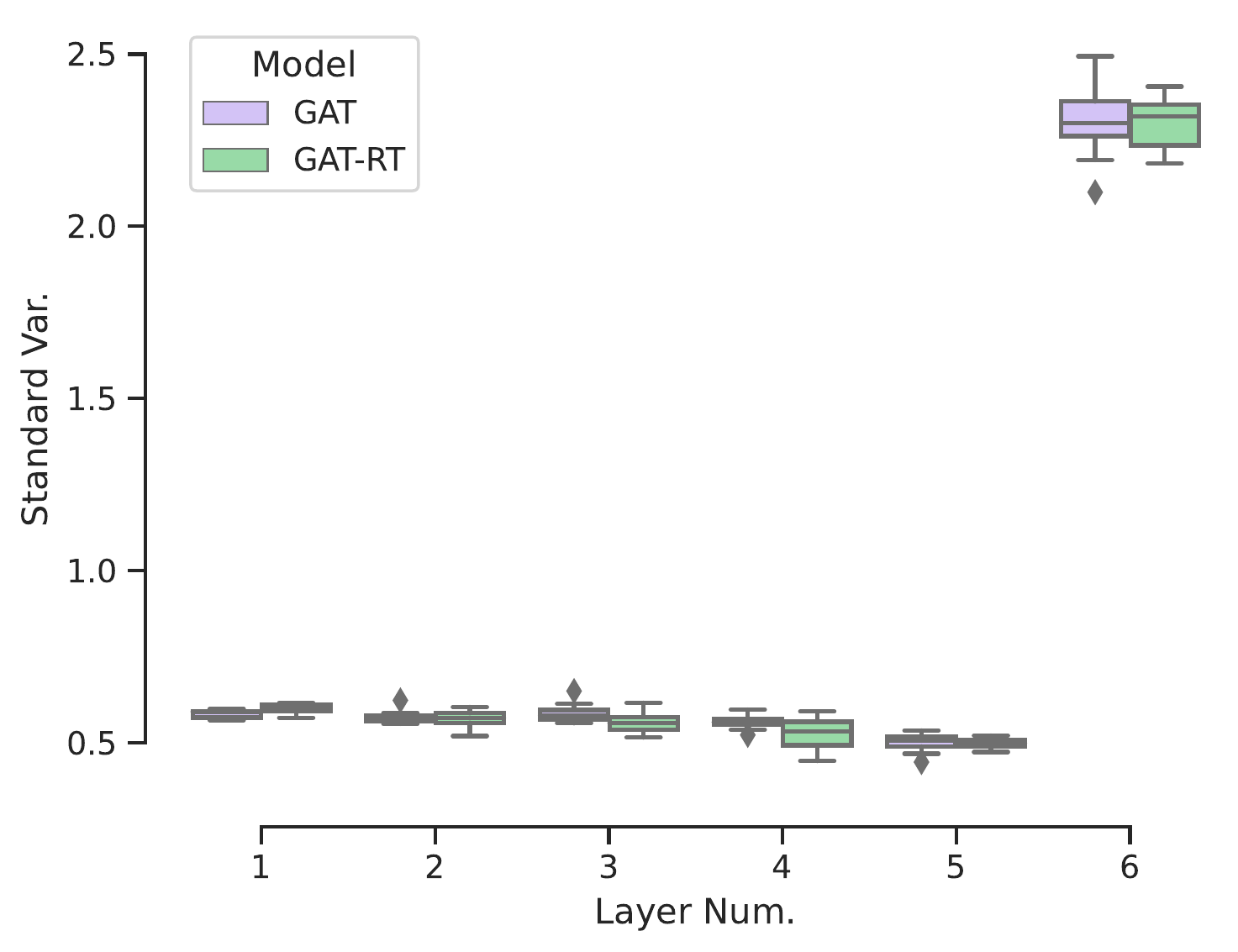}}
				\hspace{-3mm}
	\subfigure[7-layer]{
		\includegraphics[width=0.175\linewidth]{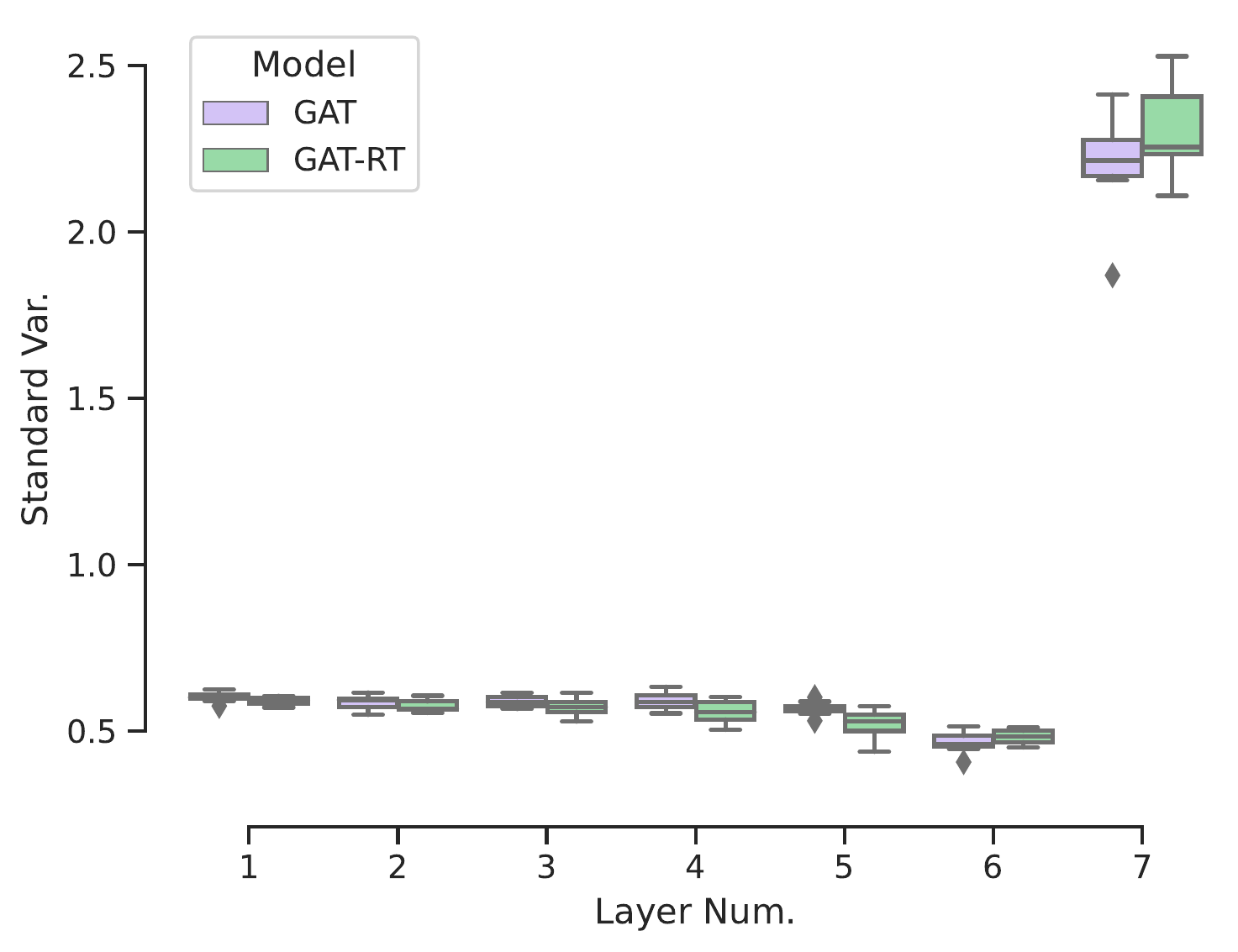}}
	\caption{Measurement of over-smoothing of GAT on ogbn-arxiv.}
	\label{Fig:distance on ogbn-arxiv}
\end{figure*}

\begin{figure*}[h]
	\centering  
	\subfigcapskip=-2pt 
	\subfigure[3-layer]{
		\includegraphics[width=0.175\linewidth]{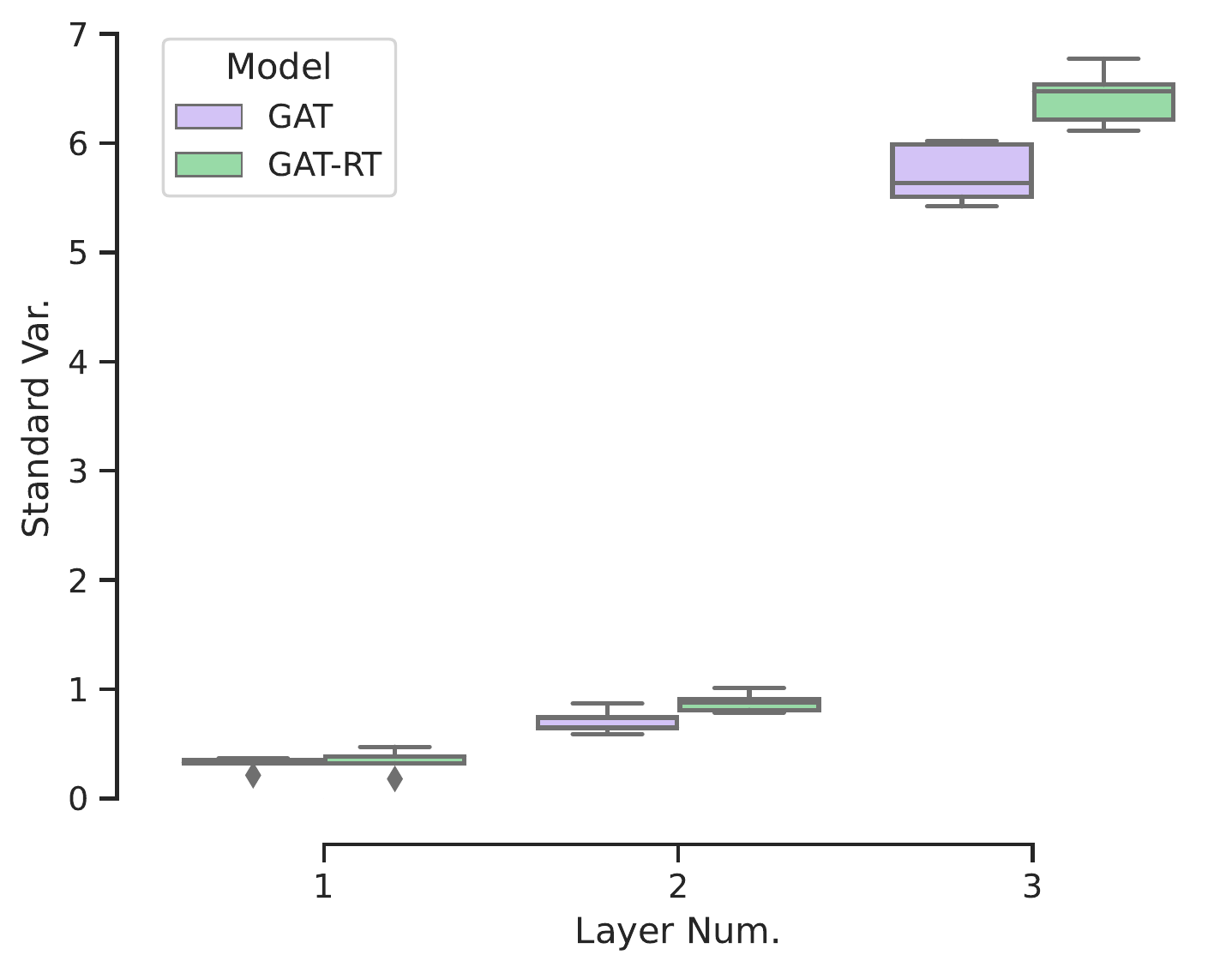}}
		\hspace{-3mm}
	\subfigure[4-layer]{
		\includegraphics[width=0.175\linewidth]{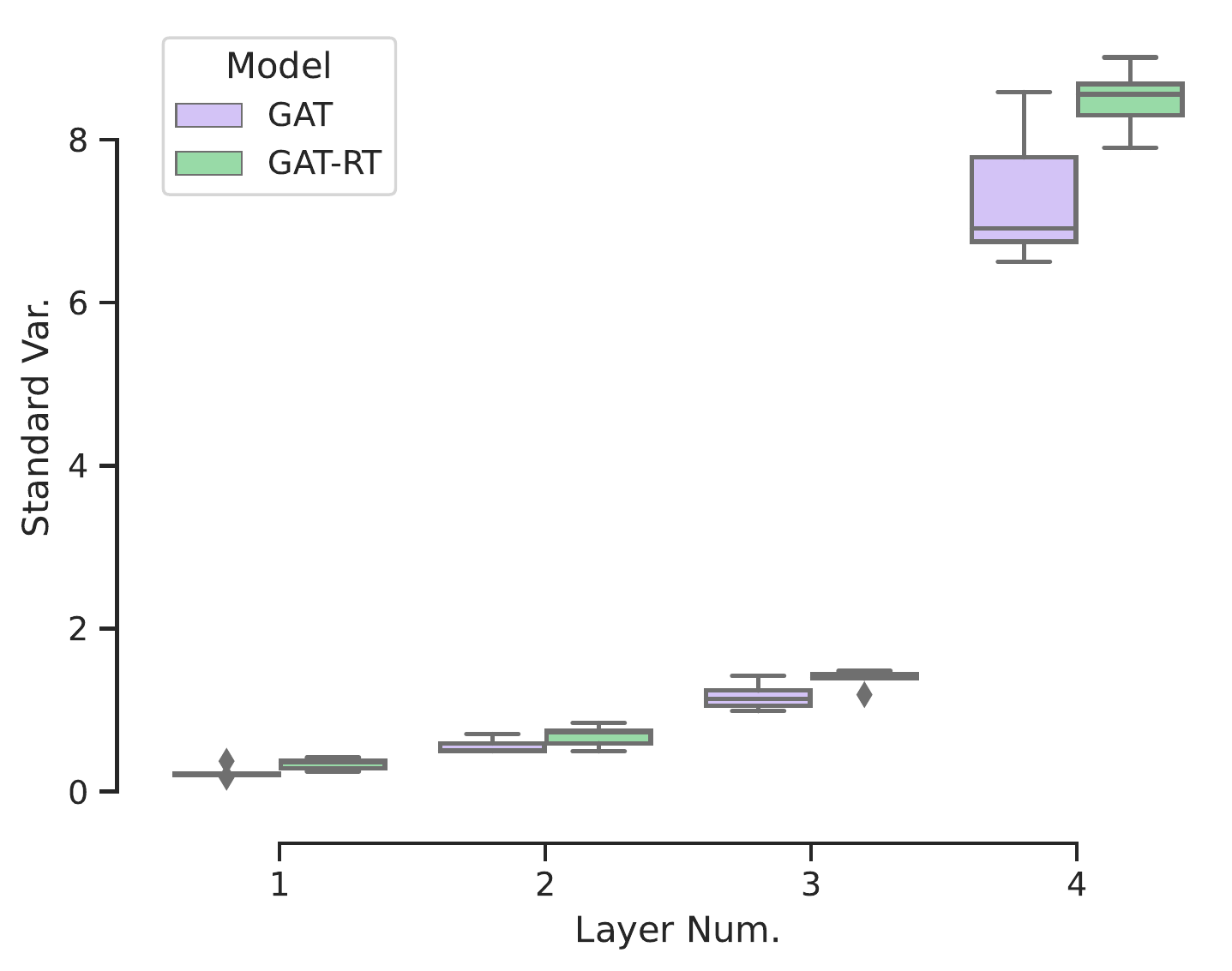}}
				\hspace{-3mm}
	\subfigure[5-layer]{
		\includegraphics[width=0.175\linewidth]{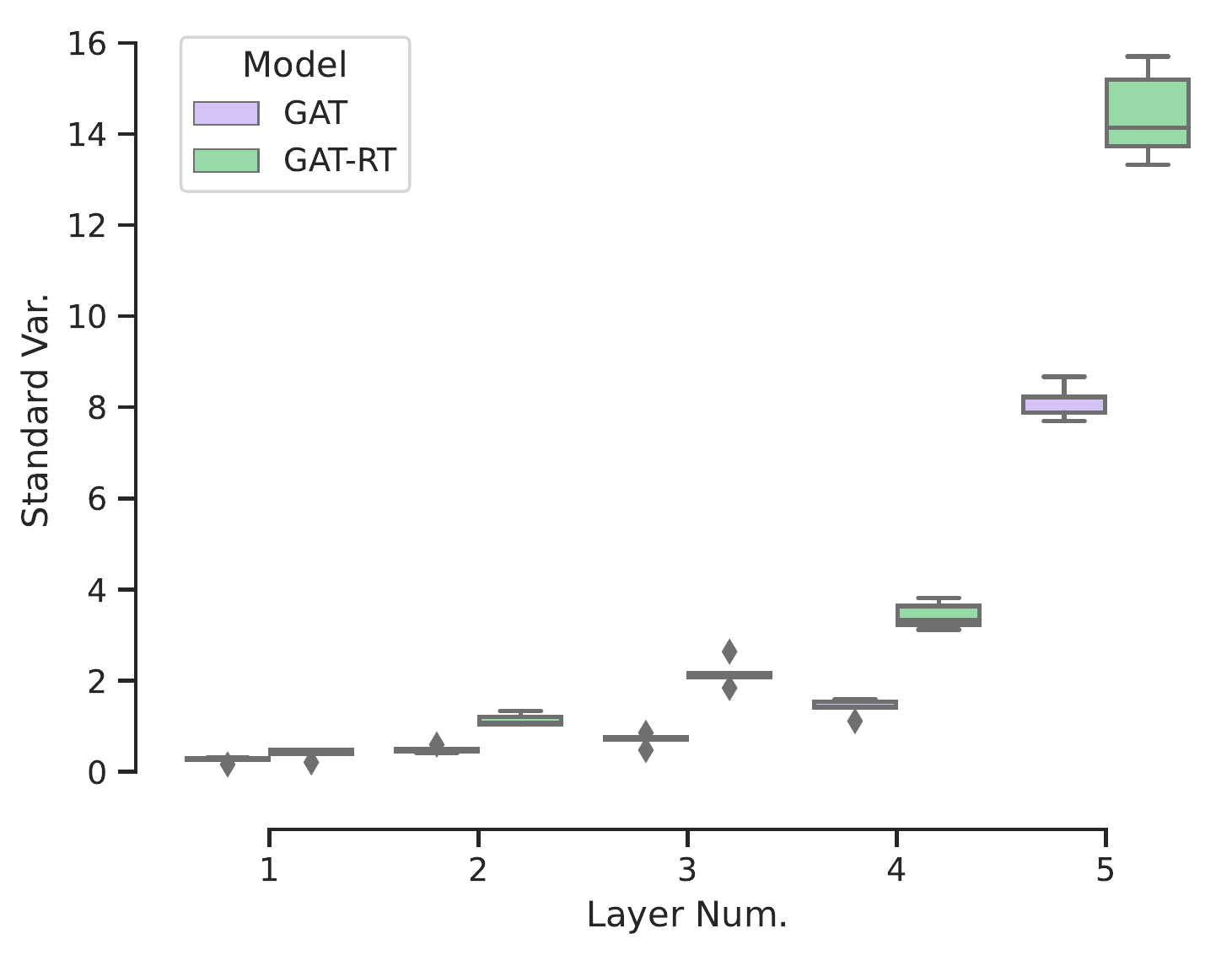}}
				\hspace{-3mm}
	\subfigure[6-layer]{
		\includegraphics[width=0.175\linewidth]{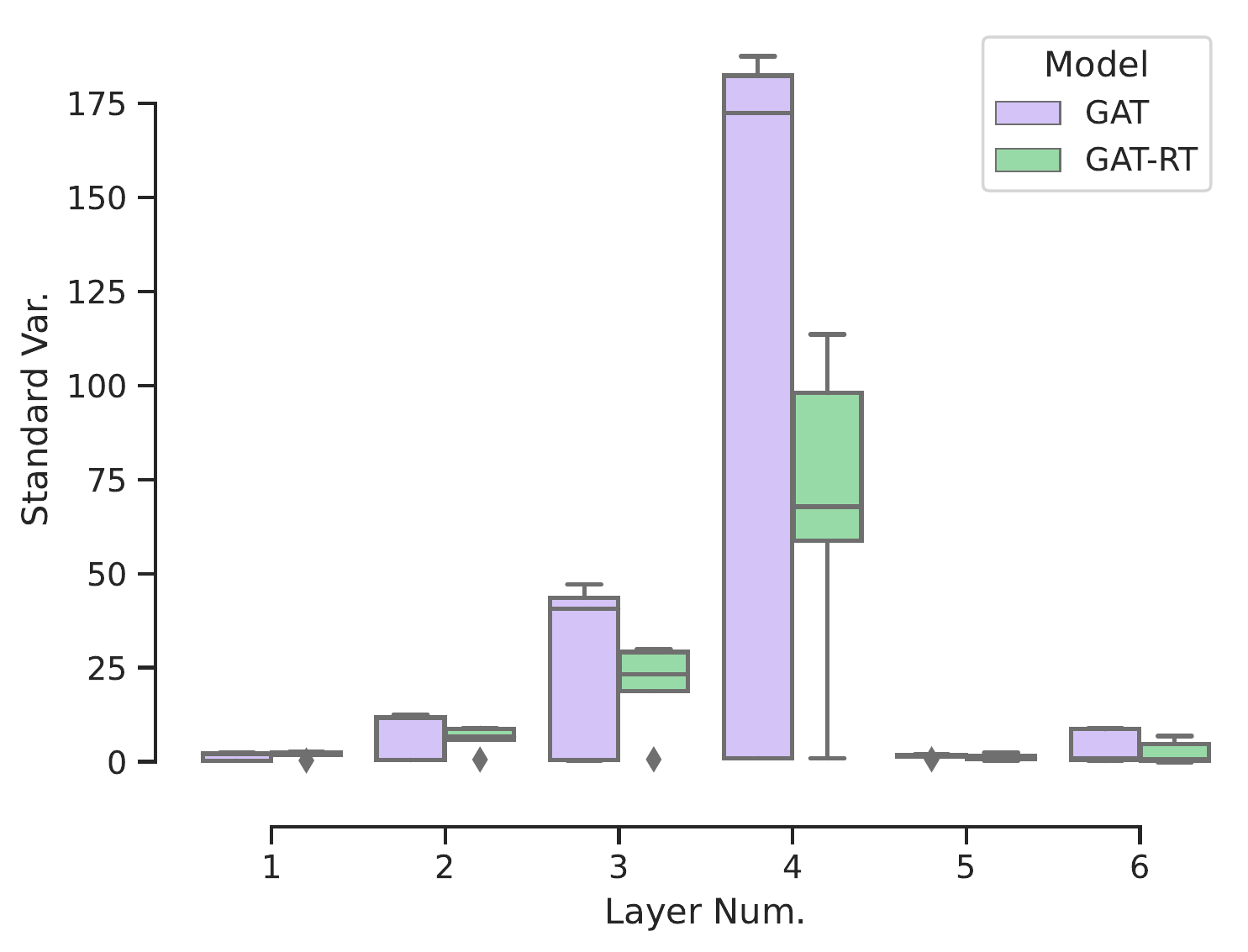}}
				\hspace{-3mm}
	\subfigure[7-layer]{
		\includegraphics[width=0.175\linewidth]{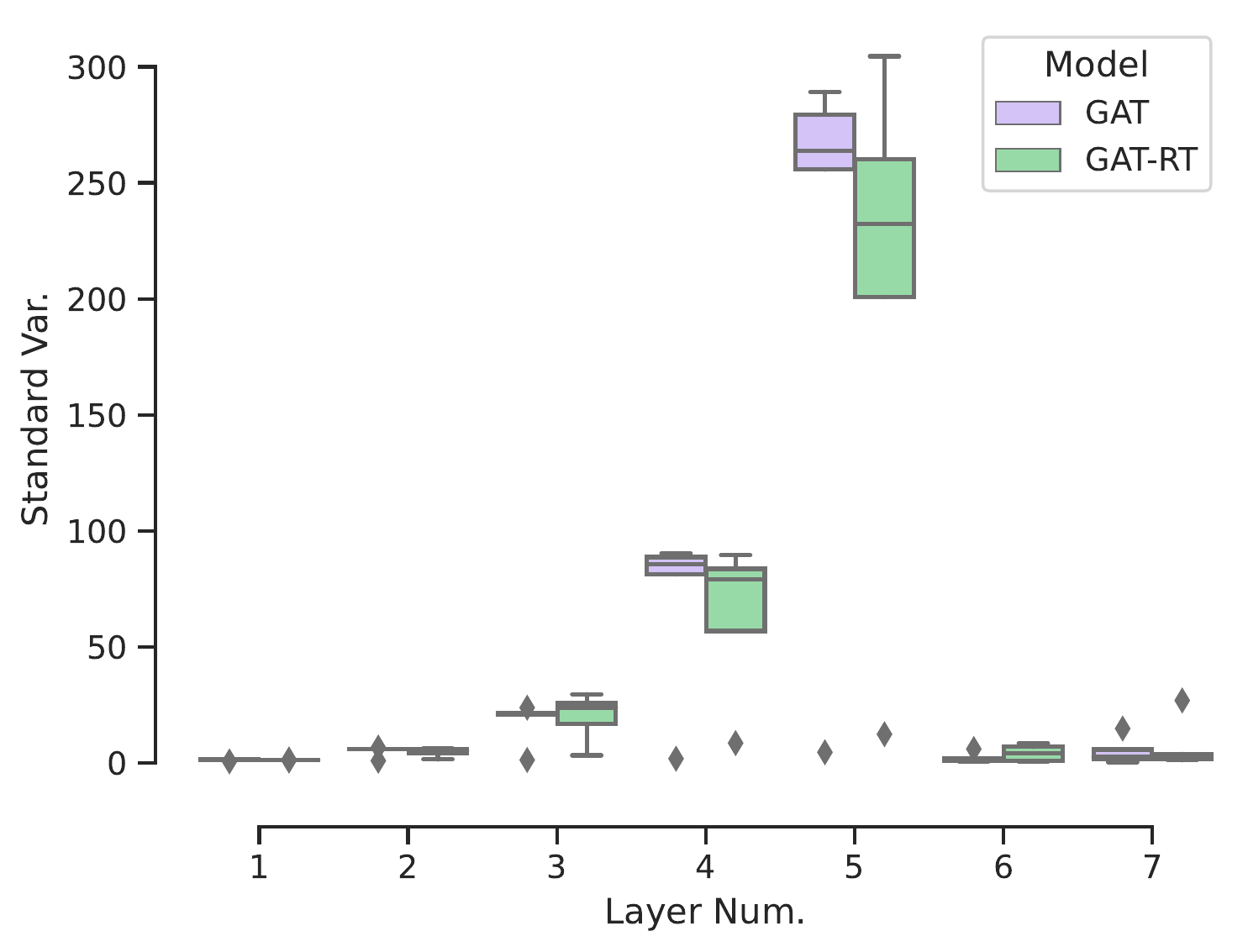}}
	\caption{Measurement of over-smoothing of GAT on Cora.}
	\label{Fig:distance on Cora}
\end{figure*}

\begin{figure*}[h]
	\centering  
	\subfigcapskip=-2pt 
	\subfigure[3-layer]{
		\includegraphics[width=0.175\linewidth]{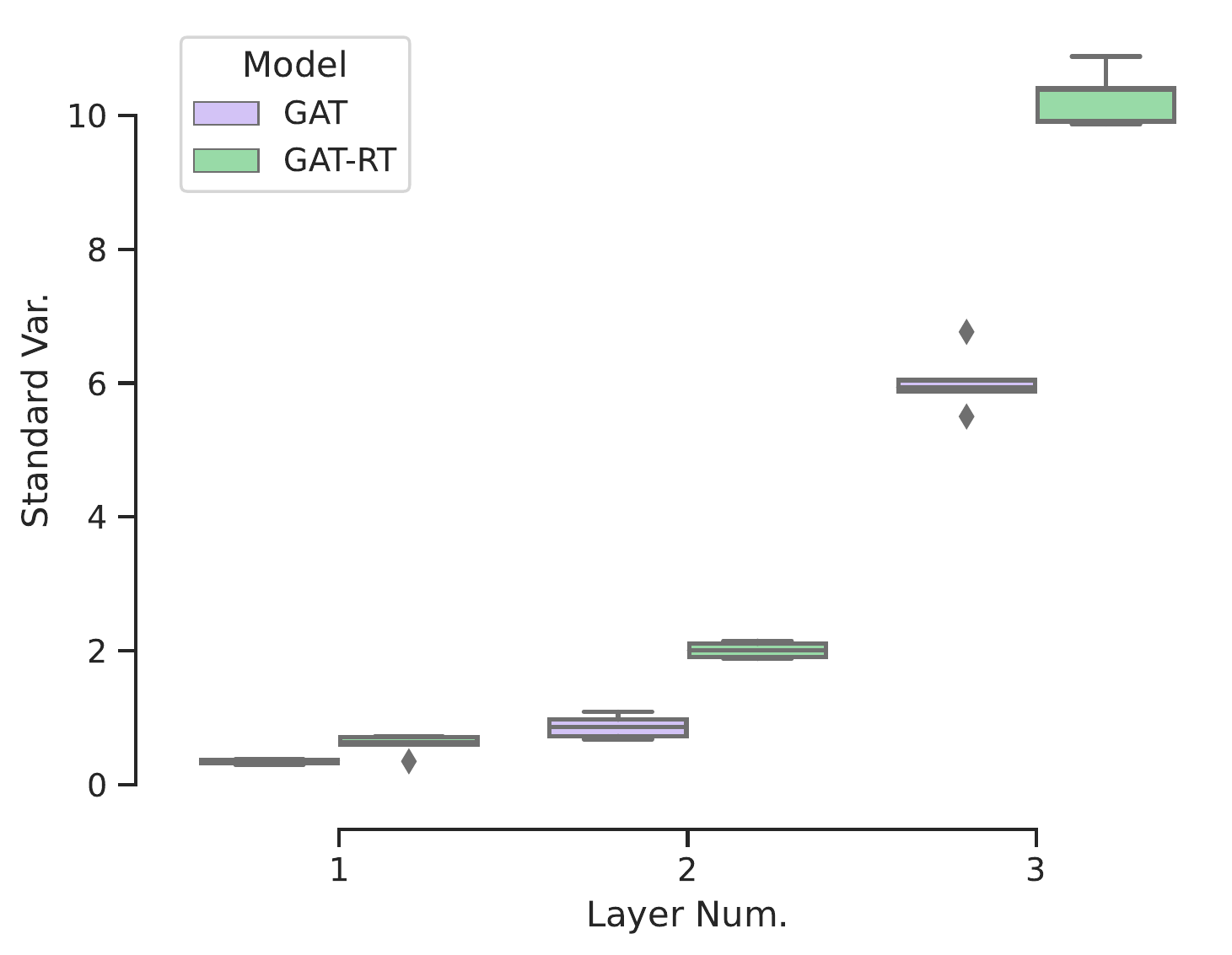}}
		\hspace{-3mm}
	\subfigure[4-layer]{
		\includegraphics[width=0.175\linewidth]{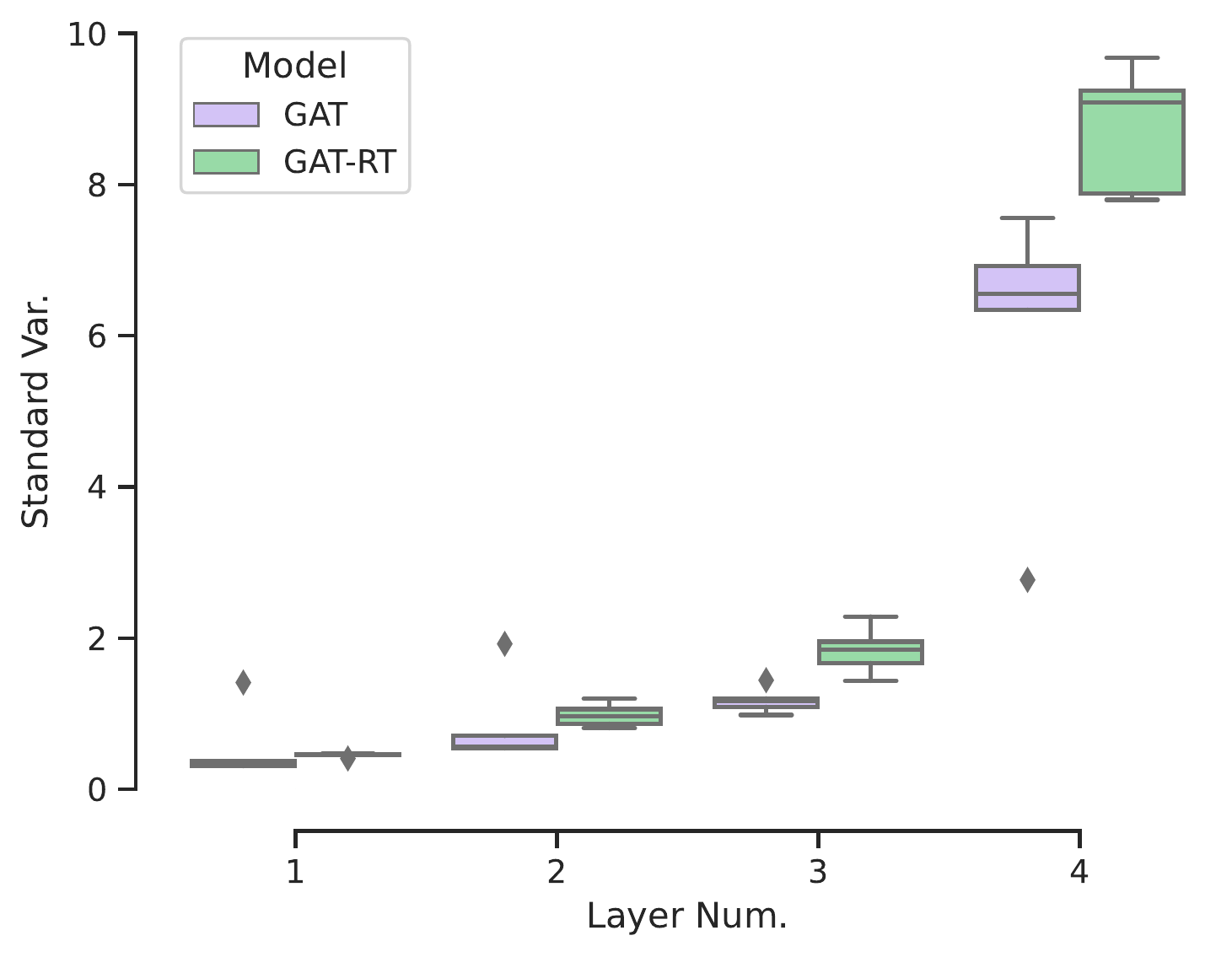}}
				\hspace{-3mm}
	\subfigure[5-layer]{
		\includegraphics[width=0.175\linewidth]{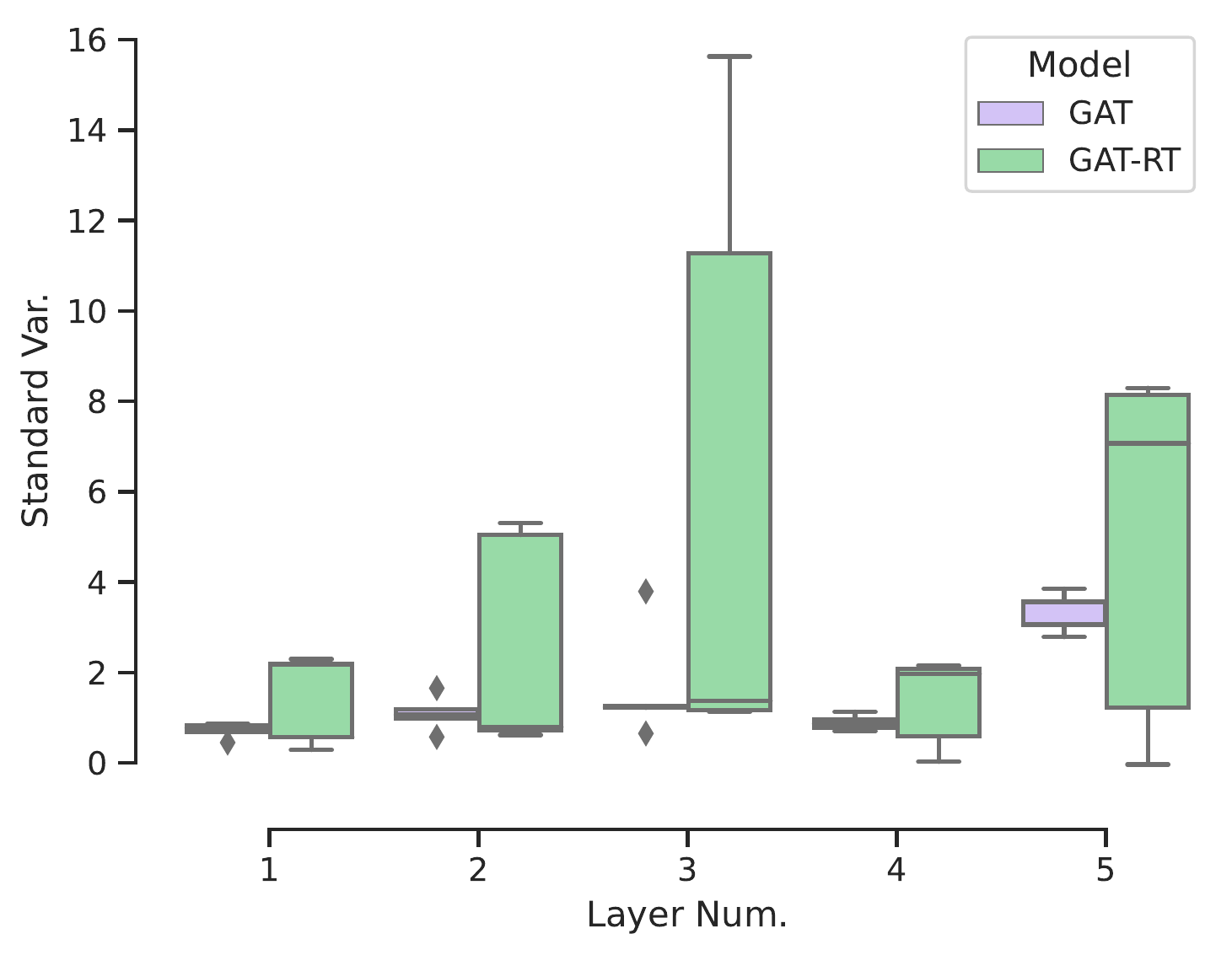}}
				\hspace{-3mm}
	\subfigure[6-layer]{
		\includegraphics[width=0.175\linewidth]{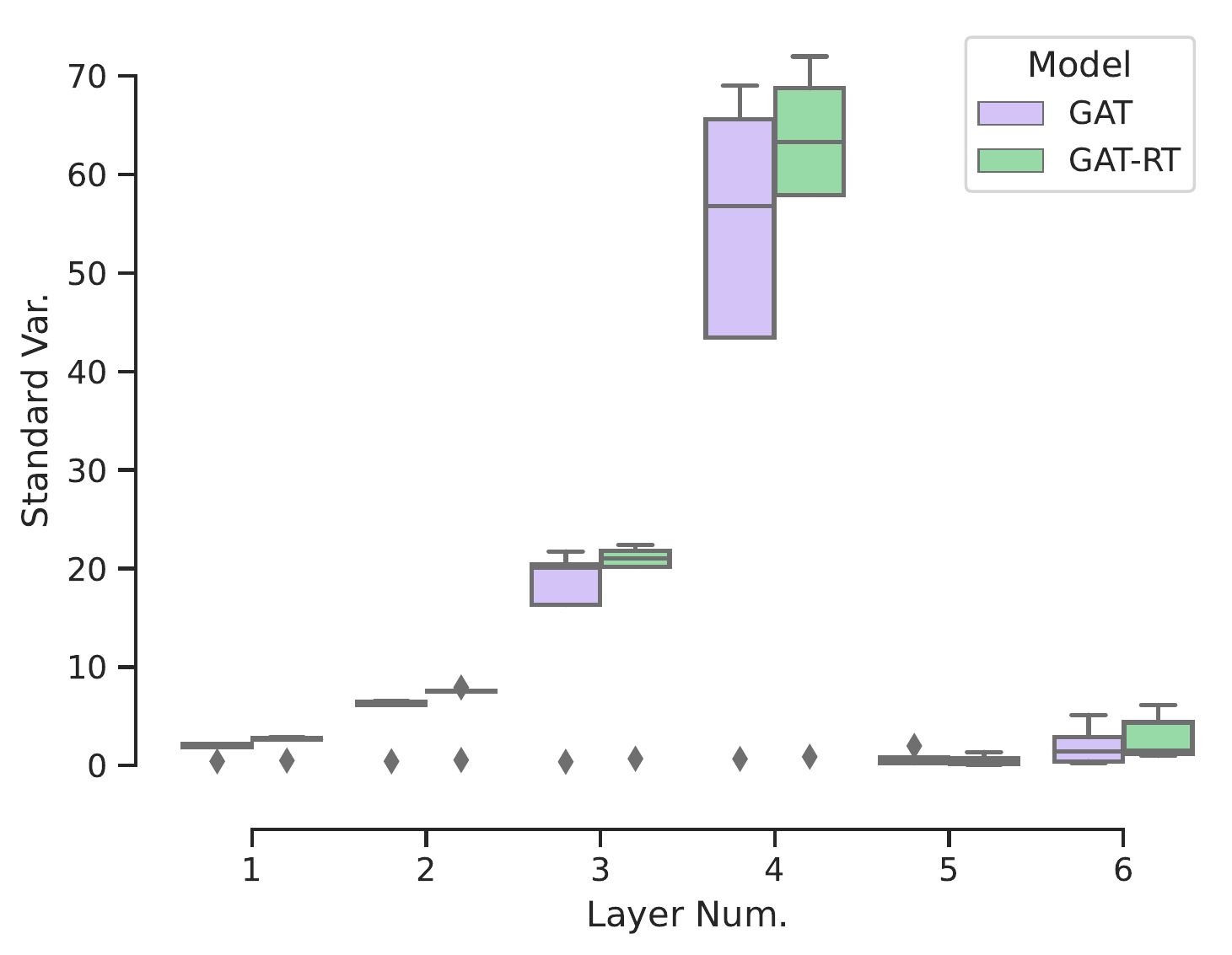}}
				\hspace{-3mm}
	\subfigure[7-layer]{
		\includegraphics[width=0.175\linewidth]{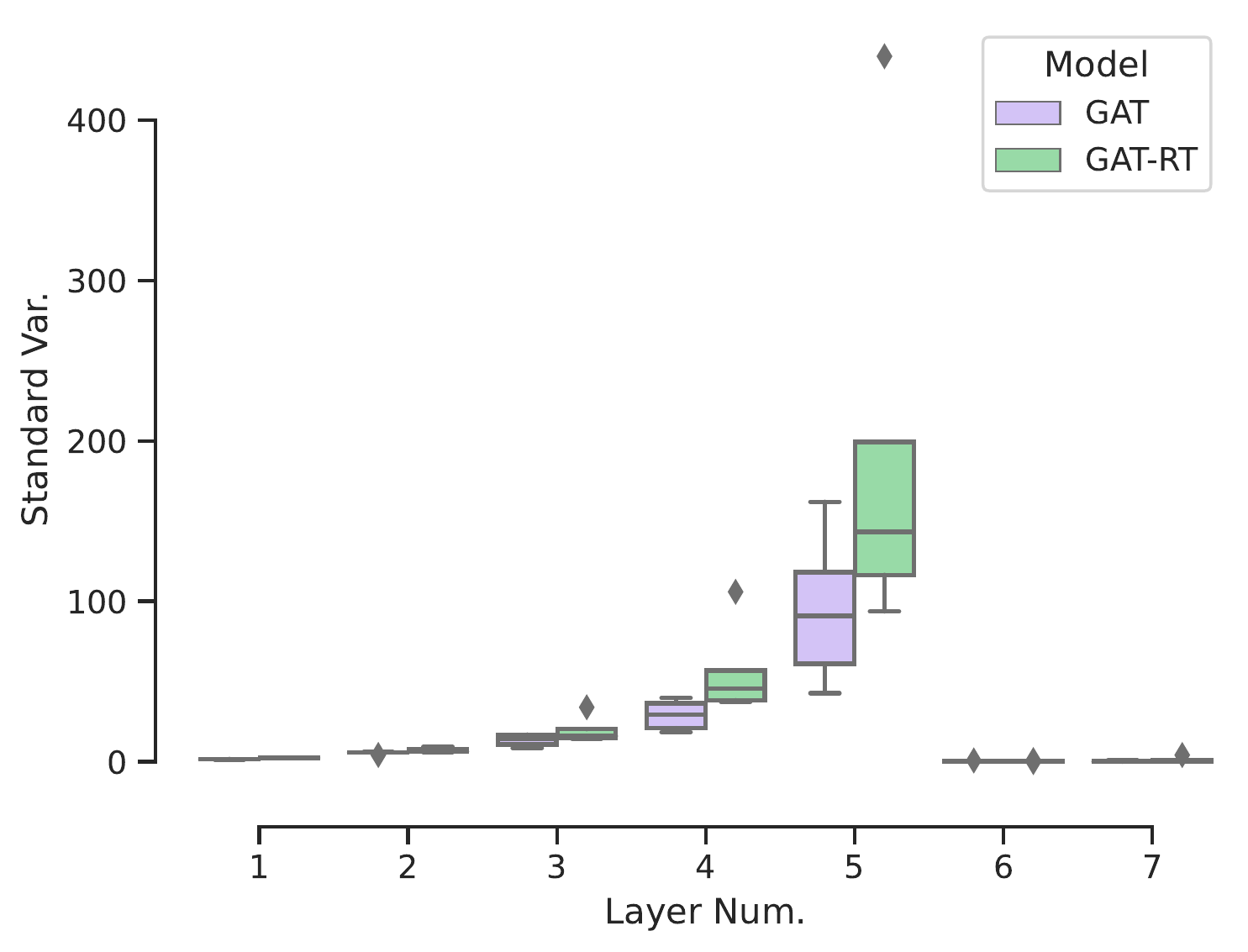}}
	\caption{Measurement of over-smoothing of GAT on Citeseer.}
	\label{Fig:distance on Citeseer}
\end{figure*}

\begin{figure*}[h]
	\centering  
	\subfigcapskip=-2pt 
	\subfigure[3-layer]{
		\includegraphics[width=0.175\linewidth]{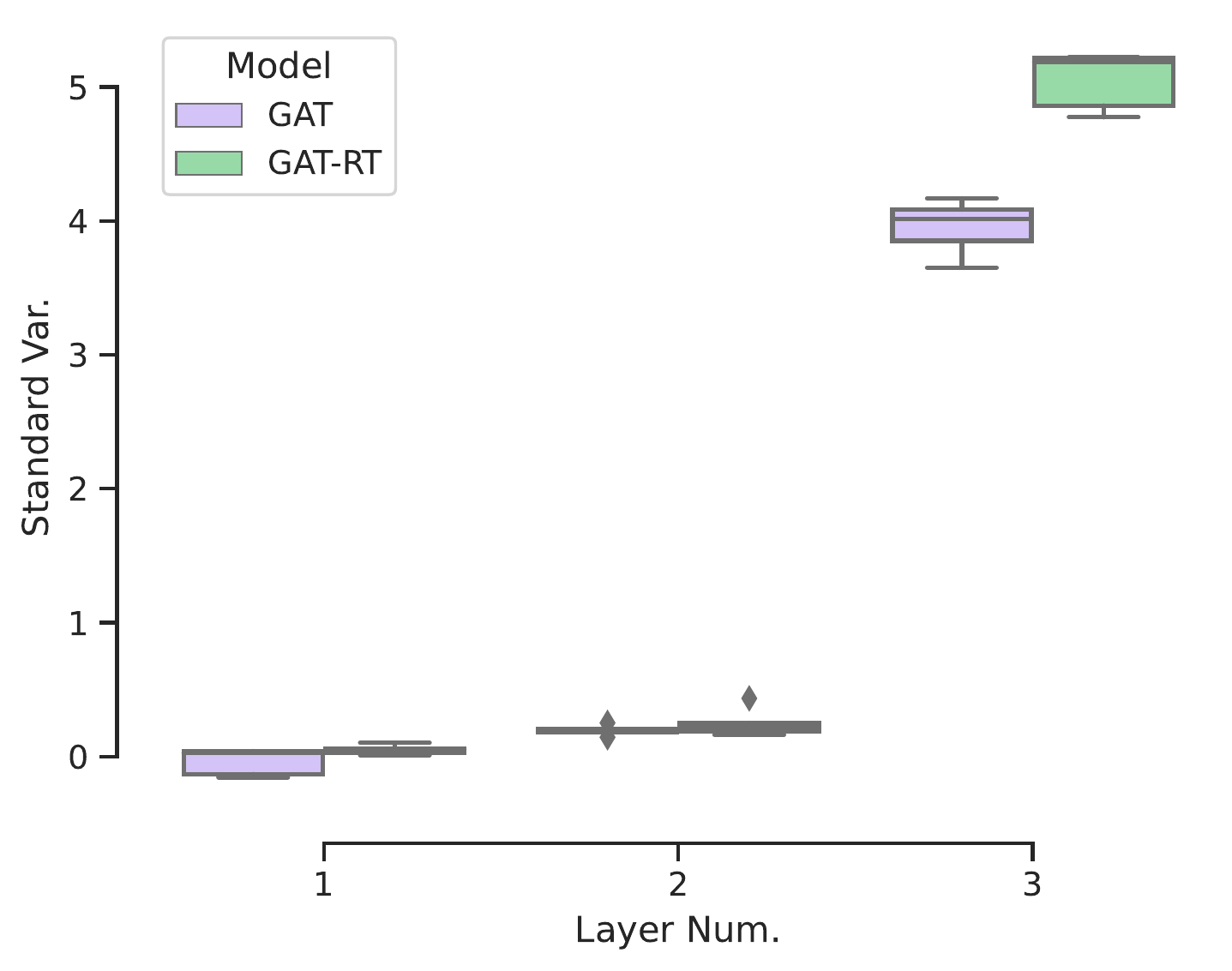}}
		\hspace{-3mm}
	\subfigure[4-layer]{
		\includegraphics[width=0.175\linewidth]{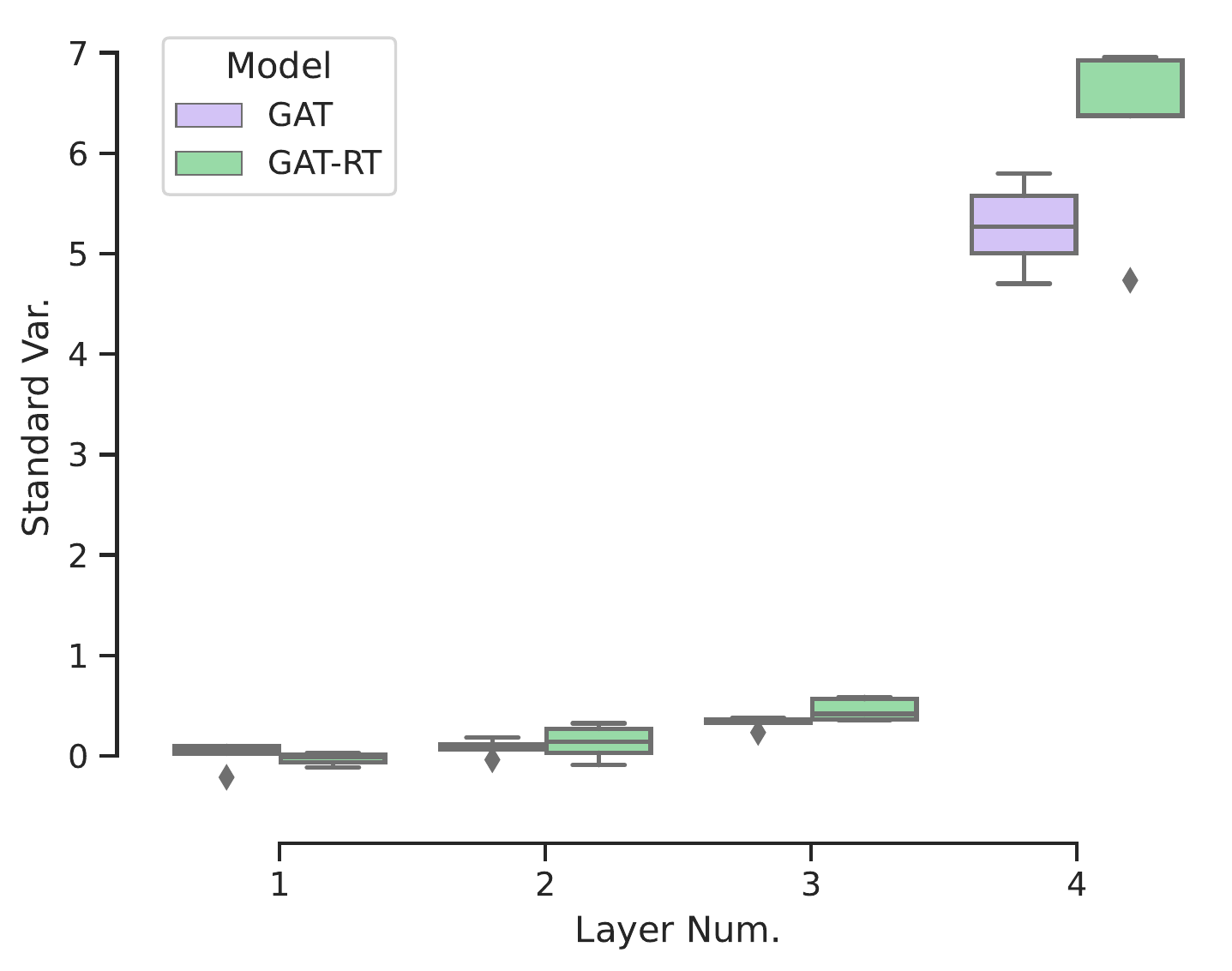}}
				\hspace{-3mm}
	\subfigure[5-layer]{
		\includegraphics[width=0.175\linewidth]{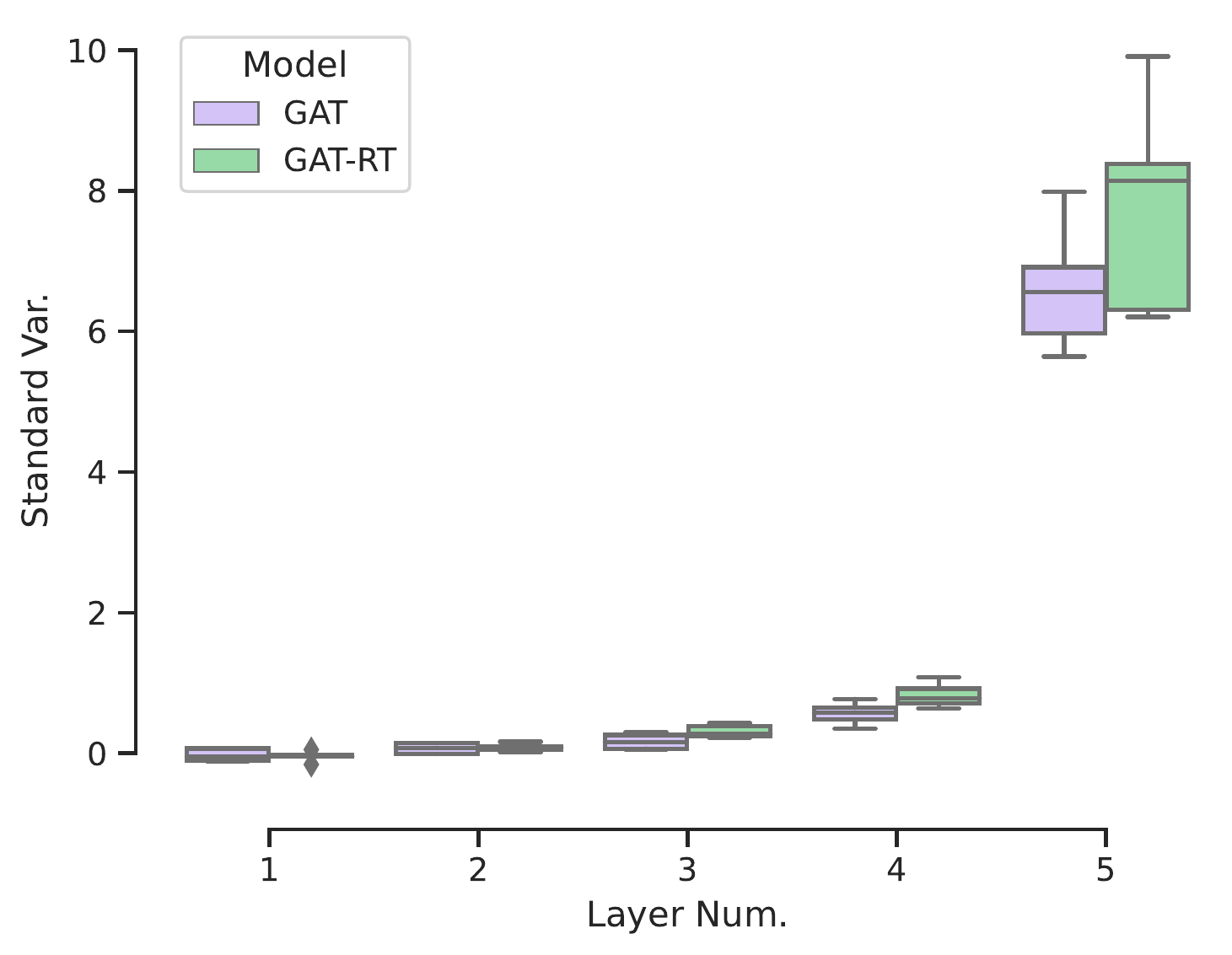}}
				\hspace{-3mm}
	\subfigure[6-layer]{
		\includegraphics[width=0.175\linewidth]{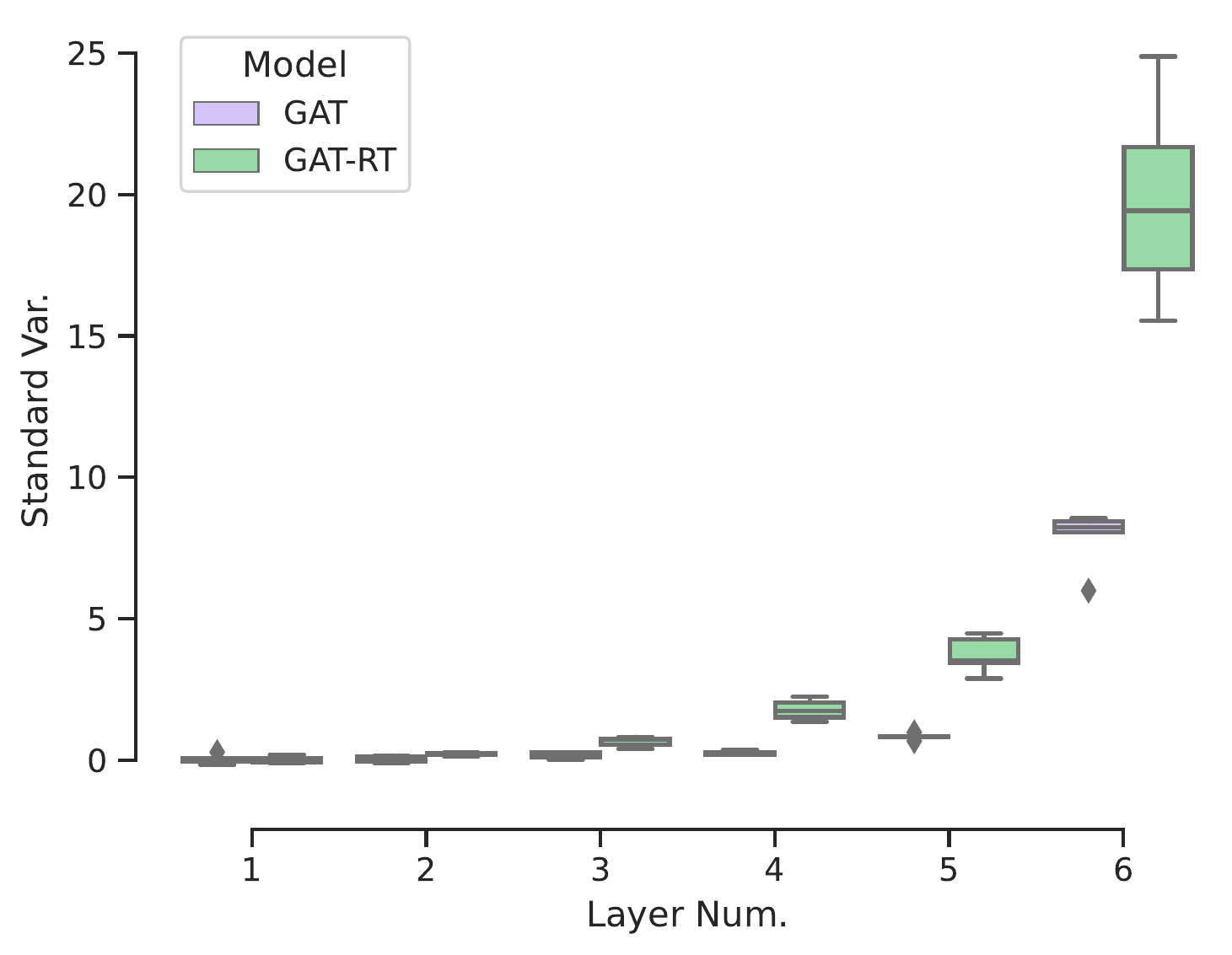}}
				\hspace{-3mm}
	\subfigure[7-layer]{
		\includegraphics[width=0.175\linewidth]{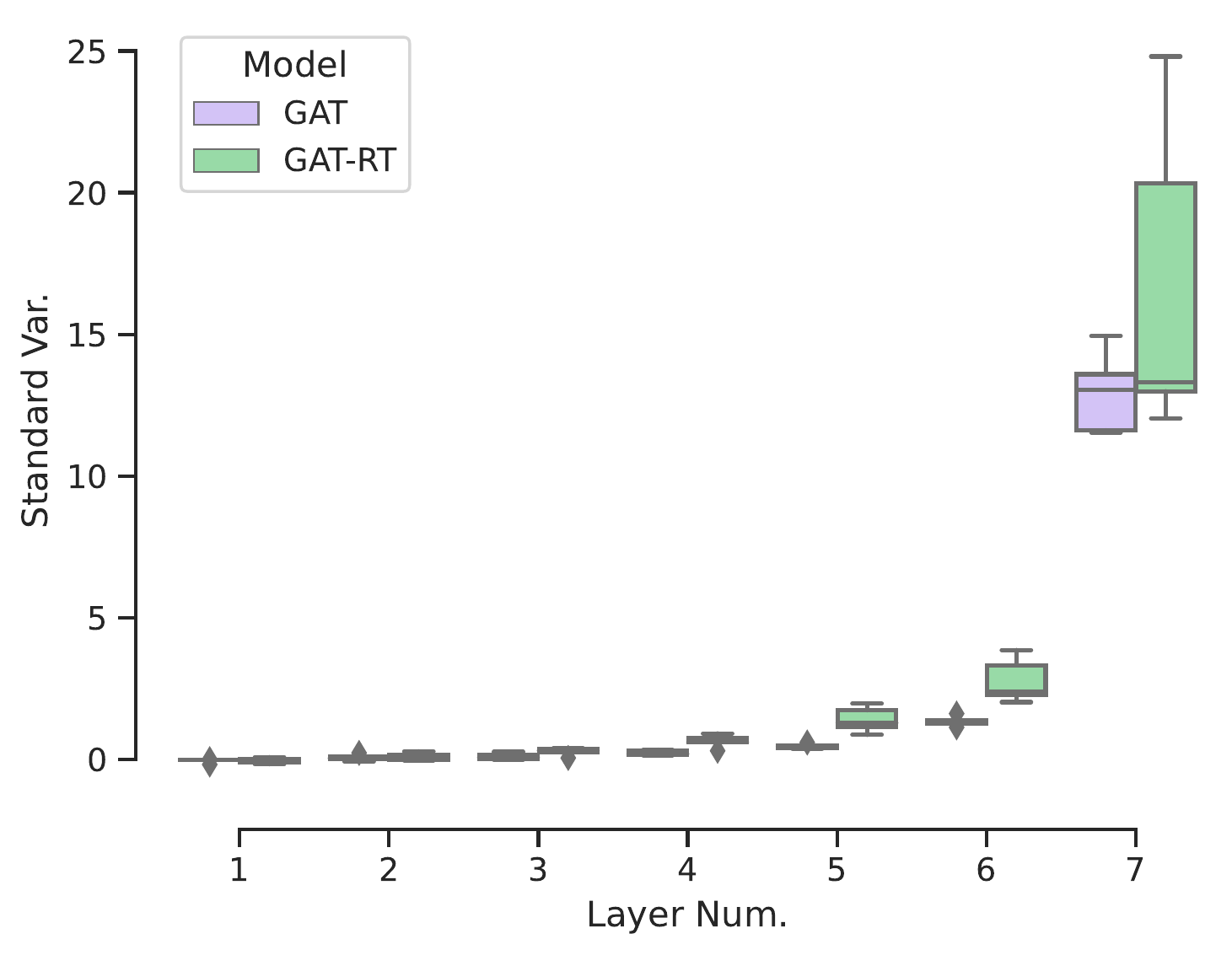}}
	\caption{Measurement of over-smoothing of GAT on Pubmed.}
	\label{Fig:distance on Pubmed}
\end{figure*}
In this subsection, we further experimentally show that the sufficient conditions in \ref{4.4} not only improve the performance of the model but also do avoid the over-smoothing problem. 

Since the neural network is a black-box model, we cannot explicitly compute the stationary distribution of the graph neural network when it is over-smoothed. Therefore we measure the degree of over-smoothing by calculating the standard deviation of each node's representation at each layer. A lower value implies more severe over-smoothing. 

Results shown in Fig.~\ref{Fig:distance on ogbn-arxiv}-\ref{Fig:distance on Pubmed} demonstrate that the node representations obtained from GAT-RT are more diverse than those from GAT, which means the alleviation of over-smoothing. It's also interesting that there is an accordance between the performance and over-smoothing, for example on Cora dataset, the performance would have a huge decrease when the number of layers is larger than 5, Fig.~\ref{Fig:distance on Cora} shows the over-smoothing phenomenon is severe at the same time. Also on Pubmed dataset, the performance is relatively stable and the corresponding Fig.~\ref{Fig:distance on Pubmed} shows that the model trained on this dataset suffers from over-smoothing lightly. These results enlighten us that over-smoothing may be caused by various objective reasons, e.g. the property of the dataset, and GAT-RT can relieve this negative effect to some extent. 

\subsection{Results of GEN}\label{sec5.4}
Because GEN \cite{li2020deepergcn} shares the same time-inhomogeneous property compared with GAT, we can obtain the similar sufficient condition using Corollary \ref{cor5.1}. See Appendix B for a detailed analysis. We conduct experiments of GEN on OGB~\cite{hu2020ogb} dataset. The detailed experimental setup is shown in Appendix C. In Table.~\ref{table:gen}, results show that there is a significant improvement in each dataset and each layer compared with the original model when adding our proposed regularization term. Due to the various tricks during the implementation of GEN such as residual connections, which may alleviate over-smoothing, the difference in the degree of over-smoothing between finite layers GEN and GEN-RT is not significant enough. We therefore don't demonstrate the degree of over-smoothing here.

    \section{Conclusion and Discussion}
		This article provides a theoretical tool for explaining and analyzing the message passing of GNNs. By establishing a connection between GNNs and Markov chains on graphs, we comprehensively study the over-smoothing problem. We reveal that the cause of over-smoothing is the convergence of the node representation distribution to stationary distribution. In our new framework, we show that although the previously proposed methods can alleviate over-smoothing, they cannot avoid over-smoothing. Through the analysis of the time-homogeneous Markv chain, we show that the operator-consistent GNN cannot avoid over-smoothing at an exponential rate. Further, we study the stationary distribution in limit sense of the general time-inhomogeneous Markov chain, and propose a necessary condition for the existence of the limiting distribution. Based on this result, we derive a sufficient condition for operator-inconsistent GNN to avoid over-smoothing in the Markovian sense. Finally, in our experiments we design a regularization term which can be flexibly added to the training. Results on the benchmark datasets show that our theoretical analysis is correct.
		
		The over-smoothing problem is still a rooted problem in message-passing based GNNs. Although we have comprehensively studied the over-smoothing problem caused by the message-passing, there are still other potential factors causing over-smoothing such as different data sets (Table \ref{table: result of gat}, \ref{table:gen}), nonlinear activation functions, etc., which need further study. In future work, GNN models that break through the message passing mechanism will be able to solve the over-smoothing problem fundamentally.
    \section*{Appendix}
	\subsection*{Proofs}
		\noindent{\bf Proof of Theorem \ref{thm4.2}}. Let the distribution of $ \xi_{1} $ be $ \mu:=\mathbf{P}\circ(\xi_{1})^{-1} $, then the distribution of $ \vec{\xi} $ is $ \nu=\mu^{\mathbb{Z}^{+}} $. Given any path $ (V_{0}=v_{0},V_{1}=v_{1},\ldots,V_{l}=v_{l}),\ v_{i}\in\mathcal{V},\ i=0,1,\ldots,l $ of $ (\vec{V},\vec{\xi}\;) $,
		$$ \begin{aligned}
			&\mathbf{P}(V_{0}=v_{0},\ldots,V_{l}=v_{l})\\&=\int_{\vec{\Xi}}\delta_{v,v_{0}}\; p(\Theta^{(1)};v_{0},v_{1})\; \cdots p(\Theta^{(l)};v_{l-1},v_{l})\;\nu(d\vec{\Xi})\\
			&=\delta_{v,v_{0}}\int_{\Xi}p(\Theta^{(1)};v_{0},v_{1})\mu(d\Theta^{(1)})\int_{\Xi}p(\Theta^{(2)};v_{1},v_{2})\mu(d\Theta^{(2)})\\&\cdots \int_{\Xi}p(\Theta^{(l)};v_{l-1},v_{l})\;\mu(d\Theta^{(l)})\\
			&=\delta_{v,v_{0}}\;\mathbf{E}\left[\frac{\theta^{(1)}(v_{0},v_{1})}{\zeta^{(1)}_{v_{0}}}\right]\cdots\mathbf{E}\left[\frac{\theta^{(l)}(v_{l-1},v_{l})}{\zeta^{(l)}_{v_{l-1}}}\right],\\
		\end{aligned} $$
		where the notations are all defined in Section \ref{3.5}, and the random variables $ \zeta^{(l)}_{v_{l-1}} $ are defined as equation (7). Since $ \Theta^{(0)},\Theta^{(1)},\ldots,\Theta^{(l)},\ldots $ are independently identical distribution, $ \vec{V} $ is a time-homogeneous Markov chain with the transition matrix $ \mathbf{E}[P(\Theta^{(1)})]=\cdots=\mathbf{E}[P(\Theta^{(l)})]=\cdots. $
		
		The following computes the transition matrix $ \mathbf{E}[P(\Theta^{(l)})] $, i.e., the expectation of $ \frac{\theta^{(l)}(u,v)}{\zeta^{(l)}_{u}} $, for all $ (u,v)\in\mathcal{E} $. Since
		$ \zeta^{(l)}_{u}\sim B(\deg(u),1-\frac{1}{|\mathcal{E}|}), $
		for $ k=1,2,\ldots,\deg(u) $,
		$$\begin{aligned}
			&\mathbf{P}\left(\frac{\theta^{(l)}(u,v)}{\zeta^{(l)}_{u}}=\frac{1}{k}\right)=\mathbf{P}(\theta^{(l)}(u,v)=1,\zeta^{(l)}_{u}=k)\\
			&=\mathbf{P}(\theta^{(l)}(u,v)=1)\mathbf{P}(\zeta^{(l)}_{u}=k|\theta^{(l)}(u,v)=1)\\
			&=\mathbf{P}(\theta^{(l)}(u,v)=1)\mathbf{P}(\zeta^{(l)}_{u}-\theta^{(l)}(u,v)=k-1|\theta^{(l)}(u,v)=1)\\
			&=(1-\frac{1}{|\mathcal{E}|})C_{\deg(u)-1}^{k-1}\left(1-\frac{1}{|\mathcal{E}|}\right)^{k-1}\left(\frac{1}{|\mathcal{E}|}\right)^{(\deg(u)-1)-(k-1)}\\
			&=C_{\deg(u)-1}^{k-1}\left(1-\frac{1}{|\mathcal{E}|}\right)^{k}\left(\frac{1}{|\mathcal{E}|}\right)^{\deg(u)-k};
				\end{aligned} $$
			$$\begin{aligned}
			&\mathbf{P}(\frac{\theta^{(l)}(u,v)}{\zeta^{(l)}_{u}}=0)=1-\sum_{k=1}^{\deg(u)}\mathbf{P}(\frac{\theta^{(l)}(u,v)}{\zeta^{(l)}_{u}}=\frac{1}{k}),
		\end{aligned} $$
		where $ C_{n}^{m}:=\frac{n!}{m!(n-m)!} $ denotes the combinatorial number, satisfying $ \frac{1}{n}\cdot C_{n}^{m}=\frac{1}{m}\cdot C_{n-1}^{m-1}. $
		The probability distribution of $\frac{\theta^{(l)}(u,v)}{\zeta^{(l)}_{u}} $ is obtained, and its expectation is calculated below
		$$ \begin{aligned}
			&\mathbf{E}\left[\dfrac{\theta^{(l)}(u,v)}{\zeta^{(l)}_{u}}\right]=\sum_{k=1}^{\deg(u)}C_{\deg(u)-1}^{k-1}\left(1-\frac{1}{|\mathcal{E}|}\right)^{k}\left(\frac{1}{|\mathcal{E}|}\right)^{\deg(u)-k}\cdot\frac{1}{k}\\
			&=\sum_{k=1}^{\deg(u)}\left(1-\frac{1}{|\mathcal{E}|}\right)^{k}\left(\frac{1}{|\mathcal{E}|}\right)^{\deg(u)-k}\cdot\frac{1}{k}\cdot C_{\deg(u)-1}^{k-1}\\
			&=\sum_{k=1}^{\deg(u)}\left(1-\frac{1}{|\mathcal{E}|}\right)^{k}\left(\frac{1}{|\mathcal{E}|}\right)^{\deg(u)-k}\cdot\frac{1}{\deg(u)}\cdot C_{\deg(u)}^{k}\\
			&=\frac{1}{\deg(u)}\cdot \left(\sum_{k=1}^{\deg(u)}C_{\deg(u)}^{k}\left(1-\frac{1}{|\mathcal{E}|}\right)^{k}\left(\frac{1}{|\mathcal{E}|}\right)^{\deg(u)-k}\right)\\
			&=\frac{1}{\deg(u)}\cdot\left(1-C_{\deg(u)}^{0}\left(\frac{1}{|\mathcal{E}|}\right)^{\deg(u)}\right)=\left(1-\frac{1}{|\mathcal{E}|^{\deg(u)}}\right)\tilde{p}_{\text{rw}}(u,v),
		\end{aligned} $$
		where $ \tilde{p}_{\text{rw}}(u,v)=\frac{1}{\deg(u)} $ is the $ u $th row, $ v $th column element of $ \tilde{P}_{\text{rw}}:=\tilde{D}^{-1}\tilde{A} $. Thus the transition matrix of the original chain $ \vec{V} $ is
		$$ P_{\text{drop}}:=\mathbf{E}[P(\Theta^{(l)})]=(I-\Gamma)\tilde{D}^{-1}\tilde{A}+\Gamma, $$
		where
		$ \Gamma:=diag\left(\frac{1}{|\mathcal{E}|^{\deg(1)}},\frac{1}{|\mathcal{E}|^{\deg(2)}},\cdots,\frac{1}{|\mathcal{E}|^{\deg(N)}}\right) $
		is a diagonal matrix. The element $ \frac{1}{|\mathcal{E}|^{\deg(u)}} $ denotes the probability that all edges connected to the node $ u $ are dropped. \hfill$\square$
		
		\noindent{\bf Proof of Theorem \ref{thm4.2.1}}. 
		(1) For $ \vec{V}_{\text{rw}} $, for all $ v\in\mathcal{V} $, since
		$$ \begin{aligned}
			&\sum_{u\in\mathcal{V}}\pi(u)\; p_{\text{rw}}(u,v)=\sum_{(u,v)\in\mathcal{E}}\frac{\deg(u)}{2|\mathcal{E}|}\frac{1}{\deg(u)}\\
			&=\frac{\deg(v)}{2|\mathcal{E}|}=\pi(v),
		\end{aligned} $$
		$ \pi $ is a stationary distribution of $ \vec{V}_{\text{rw}} $. For $ \vec{V}_{\text{lazy}} $, $ \forall v\in\mathcal{V} $, since
		$$ \begin{aligned}
			&\sum_{u\in\mathcal{V}}\pi(u)\; p_{\text{lazy}}(u,v)=\sum_{(u,v)\in\mathcal{E}}\frac{\deg(u)}{2|\mathcal{E}|}\frac{1-\gamma}{\deg(u)}+\frac{\deg(v)}{2|\mathcal{E}|}\gamma\\
			&=\frac{\deg(v)}{2|\mathcal{E}|}(1-\gamma)+\frac{\deg(v)}{2|\mathcal{E}|}\gamma=\frac{\deg(v)}{2|\mathcal{E}|}=\pi(v),
		\end{aligned} $$
		$ \pi $ also is a stationary distribution of $ \vec{V}_{\text{lazy}} $.
		~\\
		\noindent(2) Notice the intuitive definition of $ P^{l}_{\text{lazy}} $,
		$$ P^{l}_{\text{lazy}}=\sum_{i=0}^{l}C_{l}^{i}\gamma^{l-i}(1-\gamma)^{i}P_{\text{rw}}^{i}. $$
		That is, each step transition matrix $ P_{\text{lazy}} $ is an identity matrix $ I $ with independent probability $ \gamma $. And for all $ i<l $, by the Proposition \ref{prop3} we have $ \max_{u\in\mathcal{V}}\|P_{\text{rw}}^{i}(u,\cdot)-\pi\|_{TV}\ge\max_{u\in\mathcal{V}}\|P_{\text{rw}}^{l}(u,\cdot)-\pi\|_{TV}. $
		Thus
		$$ \begin{aligned}
			&\max_{u\in\mathcal{V}}\|P^{l}_{\text{lazy}}(u,\cdot)-\pi\|_{TV}\\&=\max_{u\in\mathcal{V}}\|\sum_{i=0}^{l}C_{l}^{i}\gamma^{l-i}(1-\gamma)^{i}P_{\text{rw}}^{i}(u,\cdot)-\pi\|_{TV}\\
			&=\max_{u\in\mathcal{V}}\|\sum_{i=0}^{l}C_{l}^{i}\gamma^{l-i}(1-\gamma)^{i}[P_{\text{rw}}^{i}(u,\cdot)-\pi]\|_{TV}\\
			&\ge\sum_{i=0}^{l}C_{l}^{i}\gamma^{l-i}(1-\gamma)^{i}\max_{u\in\mathcal{V}}\|P_{\text{rw}}^{l}(u,\cdot)-\pi\|_{TV}\\
			&=\max_{u\in\mathcal{V}}\|P_{\text{rw}}^{l}(u,\cdot)-\pi\|_{TV}.
		\end{aligned} $$\hfill$\square$
	
	\noindent{\bf Proof of Theorem \ref{thm4.2.2}}. 
	$$ \begin{aligned}
		P_{\text{rw}}^{l}&=(D^{-1}A)^{l}=D^{-\frac{1}{2}}(D^{-\frac{1}{2}}AD^{-\frac{1}{2}})^{l}D^{\frac{1}{2}}\\&=D^{-\frac{1}{2}}(I-L)^{l}D^{\frac{1}{2}}.
	\end{aligned} $$
	Then
	$$ \begin{aligned}
		\mu P_{\text{rw}}^{l}&=\mu D^{-\frac{1}{2}}(I-L)^{l}D^{\frac{1}{2}}=\sum_{k=1}^{N}(1-\lambda_{k})^{l}a_{k}\phi_{k}D^{\frac{1}{2}}\\
		&=\pi+\sum_{k=2}^{N}(1-\lambda_{k})^{l}a_{k}\phi_{k}D^{\frac{1}{2}},
	\end{aligned} $$
	where $ \pi=a_{1}\phi_{1}D^{\frac{1}{2}} $ is concluded from \cite{chung1997spectral}. On the other hand, the corresponding normalized Laplacian matrix of $ P_{\text{lazy}} $ is
	$$ \begin{aligned}
		L_{\text{lazy}}&=I-D_{\text{lazy}}^{-\frac{1}{2}}A_{\text{lazy}}D_{\text{lazy}}^{-\frac{1}{2}}\\
		&=I-D^{-\frac{1}{2}}((1-\gamma)A+\gamma D)D^{-\frac{1}{2}}\\
		&=I-(1-\gamma)D^{-\frac{1}{2}}AD^{-\frac{1}{2}}-\gamma I=(1-\gamma)L.
	\end{aligned} $$
	Therefore, the eigenvalue of $ L_{\text{lazy}} $ is $ 0=(1-\gamma)\lambda_{1}\leq (1-\gamma)\lambda_{2}\leq\cdots\leq(1-\gamma)\lambda_{N}< 2 $, and the eigenvector remains $ \phi_{1},\phi_{2},\cdots,\phi_{N} $. In the same way as $ \mu P_{\text{rw}}^{l} $, there are
	$$ \mu P_{\text{lazy}}^{l}=\pi+\sum_{k=2}^{N}(1-(1-\gamma)\lambda_{k})^{l}a_{k}\phi_{k}D^{\frac{1}{2}}. $$
	\hfill$\square$
	
	\noindent{\bf Proof of Theorem \ref{thm4.3.1}}. 
	Since $ \mathcal{G} $ is a connected graph and the transition matrix $ P=(p(u,v),u,v\in\mathcal{V}) $ satisfies
	$ p(u,v)>0,\ (u,v)\in\mathcal{E}, $
	for any node $ u,z\in\mathcal{V} $, there exists $ n\in\mathbb{Z}^{+} $ that satisfies
	$ p^{n}(u,z)>0. $
	Thus $ P $ is irreducible.
	
	We consider the period of any $ u\in\mathcal{V} $. Then since $ \mathcal{G} $ is a non-bipartite graph, $ \mathcal{T}(u)=1 $. Thus $ P $ is aperiodic. Then there exists a unique stationary distribution $ \pi $ of $ P $.
	\hfill$\square$
	
	\noindent{\bf Proof of Theorem \ref{thm4.3.2}}. 
	Since $ P $ is irreducible and aperiodic. Then by Lemma \ref{lem3}, there exist constants $ \alpha\in(0,1) $ and $ C>0 $ such that
	$$ \max_{u\in\mathcal{V}}\|P^{l}(u,\;\cdot\;)-\pi\|_{TV}\leq C\alpha^{n}. $$
	\hfill$\square$
	
	\noindent{\bf Proof of Theorem \ref{thm4.3.3}}. Since $ \mathcal{G} $ is a connected graph and the stochastic matrix $ P_{\text{att}}^{(l)} $ satisfies
	$ P_{\text{att}}^{(l)}(u,v)>0,\ (u,v)\in\mathcal{E}, $ for any node $ u,z\in\mathcal{V} $, there exists $ n\in\mathbb{Z}^{+} $ that satisfies
	$$ \left(P_{\text{att}}^{(l)}\right)^{n}(u,z)>0. $$
	Thus $ P_{\text{att}}^{(l)} $ is irreducible. We consider the period of any $ u\in\mathcal{V} $. Then since $ \mathcal{G} $ is a non-bipartite graph, period of $ u $ is $ 1 $. Thus $ P_{\text{att}}^{(l)} $ is aperiodic. Then there exists a unique stationary distribution $ \pi^{(l)} $ of $ P_{\text{att}}^{(l)} $.
	
	For all $ v\in\mathcal{V} $, since
	$$ \begin{aligned}
		&\sum_{u\in\mathcal{V}}\pi(u)^{(l)}P_{\text{att}}^{(l)}(u,v)\\&=\sum_{(u,v)\in\mathcal{E}}\frac{\deg^{(l)}(u)}{\sum_{k\in\mathcal{V}}\deg^{(l)}(k)}\frac{\exp(\phi^{(l)}(h_{u}^{(l-1)},h_{v}^{(l-1)}))}{\sum_{z\in\mathcal{N}(u)}\exp(\phi^{(l)}(h_{u}^{(l-1)},h_{z}^{(l-1)}))}\\
		&=\sum_{(u,v)\in\mathcal{E}}\frac{\exp(\phi^{(l)}(h_{u}^{(l-1)},h_{v}^{(l-1)}))}{\sum_{k\in\mathcal{V}}\deg^{(l)}(k)}=\frac{\deg^{(l)}(v)}{\sum_{k\in\mathcal{V}}\deg^{(l)}(k)}=\pi^{(l)}(v),
	\end{aligned} $$
	$ \pi^{(l)} $ is a stationary distribution of $ P_{\text{att}}^{(l)} $.
	\hfill$\square$\\
	
	\noindent{\bf Proof of Theorem \ref{thm5.3}}. Suppose there exists a probability measure $ \pi $ on $ E $ such that
	$ \|\mu_{n}-\pi\|\rightarrow 0, n\rightarrow \infty. $
	The following conclusion
	$$ \|\pi^{(n)}-\mu_{n-1}\|\rightarrow 0,\quad n\rightarrow \infty $$
	is proved by contradiction.
	If for any $ N\in\mathbb{N}^{+} $, there exists $ \delta>0 $, when $ n>N $, all have
	$$ \|\pi^{(n)}-\mu_{n-1}\|>\delta. $$
	Then by the triangle inequality and the Dobrushin inequality (Lemma\ref{lem2})
	$$ \begin{aligned}
		&\|\mu_{n}-\mu_{n-1}\|=\|(\pi^{(n)}-\mu_{n-1})-(\pi^{(n)}-\mu_{n})\| \\
		\ge& \|\pi^{(n)}-\mu_{n-1}\|-\|\pi^{(n)}-\mu_{n}\|\\
		=& \|\pi^{(n)}-\mu_{n-1}\|-\|\pi^{(n)}P^{(n)}-\mu_{n-1}P^{(n)}\|\\
		\ge& \|\pi^{(n)}-\mu_{n-1}\|-C(P^{(n)})\|\pi^{(n)}-\mu_{n-1}\|\\
		=&(1-C(P^{(n)}))\|\pi^{(n)}-\mu_{n-1}\|>(1-C(P^{(n)}))\delta.
	\end{aligned} $$
	And since $ C(P^{(n)})<1 $, then for any $ N\in\mathbb{N}^{+} $, there exists $ (1-C(P^{(n)}))\delta>0 $, for all $ n > N, $ $ \|\mu_{n}-\mu_{n-1}\|>(1-C(P^{(n)}))\delta. $
	By Cauchy's convergence test, it is contradictory to
	$ \|\mu_{n}-\pi\|\rightarrow 0,\quad n\rightarrow \infty. $
	Thus for any $ \epsilon>0 $, there exists $ N_{1}\in\mathbb{N}^{+} $, and when $ n>N_{1} $,
	$ \|\pi^{(n)}-\mu_{n-1}\|<\frac{\epsilon}{2}. $
	Since $ \|\mu_{n}-\pi\|\rightarrow 0,\ n\rightarrow \infty $, there exists $ N_{2}\in\mathbb{N}^{+} $, for all $ n>N_{2}, $
	$ \|\mu_{n-1}-\pi\|<\frac{\epsilon}{2}. $
	Taking $ N=\max\{N_{1},N_{2}\} $, when $ n>N $, we have
	$$ \begin{aligned}
		\|\pi^{(n)}-\pi\|&=\|(\pi^{(n)}-\mu_{n-1})+(\mu_{n-1}-\pi)\| \\
		&\leq \|\pi^{(n)}-\mu_{n-1}\|+\|\mu_{n-1}-\pi\|<\epsilon.
	\end{aligned} $$
	Then
	$ \|\pi^{(n)}-\pi\|\rightarrow 0,\quad n\rightarrow \infty.  $
	\hfill$\square$\\
	
	\noindent{\bf Proof of Corollary \ref{cor5.1}}.
	By Theorem~\ref{thm4.3.3}, for the GAT operator $ P_{\text{att}}^{(l)}, $
	$$ \pi^{(l)}(u)=\frac{\deg^{(l)}(u)}{\sum_{v\in\mathcal{V}}\deg^{(l)}(v)}\quad\forall u\in\mathcal{V}, $$
	where $ \deg^{(l)}(u):=\sum_{k\in\mathcal{N}(u)}\exp(\phi^{(l)}(h_{u}^{(l-1)},h_{k}^{(l-1)})) $ is the weighted degree of $ u $, where $$ \phi^{(l)}(h_{u}^{(l-1)},h_{k}^{(l-1)}):=\text{LeakyReLU}(\mathbf{a}^{\text{T}}[W^{(l)}h_{u}^{(l-1)}\|W^{(l)}h_{k}^{(l-1)}]). $$
	Since $ \mathcal{G} $ is connected, non-bipartite graph,
	$$ C(P_{\text{att}}^{(l)})<1. $$
	By Theorem \ref{thm5.3}, the sufficient condition for that there is no probability measure $ \pi $ on $ E $ such that
	$ \|\mu_{n}-\pi\|\rightarrow 0,\quad n\rightarrow \infty $
	is
	$$ \|\pi^{(n)}-\pi\|\nrightarrow 0,\quad n\rightarrow \infty.  $$
	
	By the Cauchy's convergence test, it is equivalent to the existence of $ \delta_{\pi}>0 $ such that for any $ l\ge1 $, satisfying
	$$ \|\pi^{(l)}-\pi^{(l+1)}\|>\delta_{\pi}. $$
	Let $ D^{(l)}:=\sum_{u\in\mathcal{V}}\deg^{(l)}(u) $, $ D_{\text{min}}=\min\{D^{(l)},D^{(l+1)}\}, $
	$$ \begin{aligned}
		&\|\pi^{(l)}-\pi^{(l+1)}\|=\sum_{u\in\mathcal{V}}\left\vert\frac{\deg^{(l)}(u)}{D^{(l)}}-\frac{\deg^{(l+1)}(u)}{D^{(l+1)}}\right\vert\\
		>&\left\vert\frac{\deg^{(l)}(u)}{D^{(l)}}-\frac{\deg^{(l+1)}(u)}{D^{(l+1)}}\right\vert>\frac{1}{D_{\text{min}}}\left\vert\deg^{(l)}(u)-\deg^{(l+1)}(u)\right\vert.
	\end{aligned} $$
	Notice that $\deg^{(l)}(u):=\sum_{k\in\mathcal{N}(u)}\exp(\text{LeakyReLU}(\mathbf{a}^{\text{T}}[W^{(l)}h_{u}^{(l-1)}\|W^{(l)}h_{k}^{(l-1)}])).$ For message passing, we default that $ \mathbf{a},W^{(l)}h_{u}^{(l-1)}\neq 0 $. Then if there exists $ \delta>0 $ such that for any $ l\ge2 $, satisfying
	$ \|h_{u}^{(l-1)}-h_{u}^{(l)}\|>\delta, $
	there must exist $ \delta_{\pi}>0 $ such that for any $ l\ge1 $, satisfying
	$$ \|\pi^{(l)}-\pi^{(l+1)}\|>\delta_{\pi}. $$
	\hfill$\square$
	\subsection*{Analysis of GEN}\label{appB}
	In this appendix, we analyze Generalized Aggregation Networks (GEN-SoftMax) \cite{li2020deepergcn}, which is a operator-inconsistent GNN model proposed for training deeper GNNs.
	
	In order to be able to train deeper GNN models, \cite{li2020deepergcn} proposes a new message passing method between nodes $ u $ and $ v $
	$$ \lambda_{u,v}^{(l)}:=\frac{\exp(\beta\mathbf{m}^{(l-1)}_{u,v})}{\sum_{k\in\mathcal{N}(u)}\exp(\beta\mathbf{m}^{(l-1)}_{u,k})}, $$
	where $ \beta $ is inverse temperature and
	$$ \mathbf{m}^{(l)}_{u,v}:=\text{ReLU}(h_{v}+\mathbb{I}(h^{(l)}_{(u,v)})\cdot h^{(l)}_{(u,v)})+\epsilon\qquad v\in\mathcal{N}(u), $$
	where $ \mathbb{I}(\;\cdot\;) $ is an indicator function being $ 1 $ when edge features exist otherwise $ 0 $, $ \epsilon $ is a small positive constant chosen as $ 10^{-7} $. Then the definition of message passing in GEN-SoftMax is
	$$ h_{u}^{(l)}:=\sum_{v\in\mathcal{N}(u)}\lambda_{u,v}^{(l)}h_{v}^{(l-1)}. $$
	Write in matrix form
	$ H^{(l)}=P_{\text{GEN}}^{(l)}H^{(l-1)}, $
	where $ P_{\text{GEN}}^{(l)}\in\mathbb{R}^{N\times N}$ satisfies $ P_{\text{GEN}}^{(l)}(u,v)=\lambda_{u,v}^{(l)} $ if $ v\in\mathcal{N}(u) $ otherwise $ P_{\text{GEN}}^{(l)}(u,v)=0 $, and $ \sum_{v=1}^{N}P_{\text{GEN}}^{(l)}(u,v)=1 $.
	
	Similar to the discussion of GAT in Section 3.3, we can relate GEN-SoftMax to a time-inhomogeneous Markov chain $ \vec{V}_{\text{GEN}} $ with a family of transition matrices of
	$$ \left\{P_{\text{GEN}}^{(1)},P_{\text{GEN}}^{(2)},\ldots,P_{\text{GEN}}^{(l)},\ldots\right\}. $$
	According to the discussion in Section \ref{4.3.2}, GEN-SoftMax does not necessarily oversmooth.
	
	Similar to Corollary 16, we have the following sufficient condition to ensure that GEN-SoftMax avoids over-smoothing.
	
	\begin{corollary}\label{corB.1}
		Let $ h_{u}^{(l)} $ be the $ l $th layer hidden layer feature on node $ u\in\mathcal{V} $ in GEN-SoftMax, then a sufficient condition for GEN-SoftMax to avoid over-smoothing in the Markovian sense is that there exists a hyperparameter $ \delta>0 $ such that for any $ l\ge1 $, satisfying
		$$ \|h_{u}^{(l)}-h_{u}^{(l+1)}\|>\delta. $$
		
	\end{corollary}
	
	\subsection*{Experimental details}\label{appC}
	In this appendix, we add more details on the experiments. Table \ref{table:dataset} shows the basic information of each dataset used in our experiments.
	
	\begin{table*}[hbpt!]
		\centering
		\caption{Summary of the statistics and data split of datasets.}
		\label{table:dataset}
		~\\
		\setlength{\tabcolsep}{5mm}{
			\begin{tabular}{l|c|c|c|c|c|c|c}
				\hline
				Dataset& (Avg.) Nodes & (Avg.) Edges & Features&Class&Train(\#/\%)&Val.(\#/\%)&Test(\#/\%)\\
				\hline
				Cora &2708(1 graph)&5429&1433&7&140&500&1000\\
				Citeseer &3327(1 graph)&4732&3703&6&120&500&1000\\
				Pubmed &19717(1 graph)&44338&500&3&60&500&1000\\
				ogbn-arvix &169,343(1 graph)&1,166,243&128&40&0.54&0.18&0.28\\
				ogbg-molhiv &25.5(41,127 graph)&27.5&9&2&0.8&0.1&0.1\\
				ogbg-ppa &243.4(158,100 graph)&2,266.1&7&37&0.49&0.29&0.22\\
				\hline
		\end{tabular}}
	\end{table*}
	Table~\ref{table:configuration} demonstrates the configuration of GNN models, actually, we keep the same setting in the corresponding paper, the only difference is we add the extra proposed regularization term in the optimization objective.
	
	\begin{table*}[hbpt!]
		\centering
		\caption{Training configuration}
		\label{table:configuration}
		~\\
		\setlength{\tabcolsep}{5mm}{
			\begin{tabular}{l|l|ccccccc}
				\hline
				Model &Dataset& Hidden. & LR.&Dropout&Epoch&Block&GCN Agg.&$\beta$\\
				\hline
				\multirow{3}{*}{GAT}
				&Cora &64&1e-2&0.5&500&-&-&-\\
				&Citeseer &64&1e-2&0.5&500&-&-&-\\
				&Pubmed &64&1e-2&0.5&500&-&-&-\\
				\hline
				
				\multirow{3}{*}{GEN}
				&ogbn-arvix &256&1e-4&0.2&300&Res+&softmax\_sg&1e-1 \\
				&ogbg-molhiv &128&1e-3&0.5&500&Res+&softmax&1\\
				&ogbg-ppa &128&1e-2&0.5&200&Res+&softmax\_sg&1e-2\\
				
				\hline
		\end{tabular}}
	\end{table*}
	In Table~\ref{table:thres}, we show the detailed selection of threshold $ T $ in equation (11). 
	\begin{table*}[hbpt!]
		\centering
		\caption{Selection of threshold $ T $ on different layer numbers}
		\label{table:thres}
		~\\
		\setlength{\tabcolsep}{5mm}{
			\begin{tabular}{l|c|cccccc}
				\hline
				\multirow{2}{*}{datasets}&\multirow{2}{*}{model}&\multicolumn{6}{c}{\#layers}\\
				\cline{3-8}
				&&3&4&5&6&7&8 \\
				\hline
				Cora&GAT-RT&1 &0.5 &0.6&0.8 &1 &0.8\\
				Citeseer&GAT-RT&0.3 &0.3 &1&0.4 &0.2 &0.1\\
				Pubmed&GAT-RT&0.3 &0.2 &0.2&0.8 &0.5 &0.4\\
				\hline
				\hline
				
				{ }& &7 &14 &28&56 \\
				\hline
				
				{ogbn-arxiv}&GEN-RT&0.3 &0.1 &(0.1)0.3&- \\
				
				{ogbg-molhiv}&GEN-RT&0.7 &0.1 &0.5&1 \\
				
				{ogbg-ppa}&GEN-RT&0.1 &0.3 &0.1&- \\
				
				\hline
				
		\end{tabular}}
	\end{table*}
	\section*{Acknowledgment}
	This paper is supported by the National Key R$\&$D Program of China project (2021YFA1000403), the National Natural Science Foundation of China (Nos. 11991022,U19B2040) and Supported by the Strategic Priority Research Program of Chinese Academy of Sciences (Grant No. XDA27000000) and the Fundamental Research Funds for the Central Universities.

\bibliographystyle{unsrtnat}
\bibliography{references}  






\end{document}